# Data Portability Revisited: Toward the Human-Centric, AI-Driven Data Ecosystems of Tomorrow

Mark Fenwick Professor of International Business Law
*Graduate School of Law, Kyushu University*

Michael Fertik Managing Partner
*Heroic Ventures*

Paulius Jurcys Senior Lecturer
*Vilnius University Law Faculty*

Timo Minssen Professor of Law
*Copenhagen University Faculty of Law*

# Data Portability Revisited: Toward the Human-Centric, AI-Driven Data Ecosystems of Tomorrow

*Mark Fenwick,*[*] *Michael Fertik,*[**] *Paulius Jurcys,*[***] *Timo Minssen*[****]

**ABSTRACT**

*This Article critically examines the contemporary regulatory framework and discourse surrounding data portability in the United States. Using recent regulatory developments in the European Union as an illustration, this Article suggests that although data access and portability are identified as vital issues in multiple policy instruments, in its current iteration, at least, legal conceptions of portability continue to reinforce the interests of service providers and data controller enterprises rather than individual end users. This Article argues that a paradigm shift toward a more human-centric data approach to data governance must occur, under which data would be recognized as fundamental to an individual's identity in a digital age. Therefore, it should be placed in the hands of individuals rather than service providers or data controller enterprises. This Article considers technical and market trends in the European Union that reveal and facilitate such a change. It suggests that regulatory frameworks should better align with these technological and market developments to encourage change-inducing trends among market actors. In short, this Article identifies a transformative approach to data portability that empowers individuals with the freedom and ability to aggregate their data in secure personal spaces under their control or dominion. Such a human-centric perspective on data portability is crucial in building Artificial Intelligence (AI)-powered applications for individual*

* Professor of International Business Law, Graduate School of Law, Kyushu University, Japan.

** Managing Partner, Heroic Ventures and the CEO and Founder of Modelcode.ai.

*** Senior Lecturer at Vilnius University Law Faculty, Co-Founder, Prifina.

**** Professor of Law and Director of the Center for Advanced Studies in Biomedical Innovation Law (CeBIL), Copenhagen University Faculty of Law.

*consumers that can pave the way for the human-centric, AI-driven data ecosystems of the future.*

**TABLE OF CONTENTS**

## I. INTRODUCTION[1]

Public surveys conducted across multiple jurisdictions around the world reveal that the overwhelming majority of individuals are concerned about the fate of their data, including *where* the data goes, *who* has access to it, and *how* it is used by data controllers and other third parties with whom such data is shared.[2] And yet, this growing sense of unease regarding personal data raises some obvious, albeit tricky, questions: if personal data is so valuable, why are people so willing to share it with others, and why do they typically choose to give it away for free to third parties (especially tech firms, notably Big Tech)[3] whom they simultaneously claim not to trust?[4]

People typically exercise a much greater level of caution when it comes to other personal possessions that have this combination of subjective and objective value. They safeguard their possessions by storing them securely in their homes or entrusting them to reliable third parties with expertise in security, like banks. So, why is the

1. This publication could only consider developments that have occurred until September 2024. Note in particular that the EHDS Regulation entered into force on 26 March 2025 and will become applicable in different phases according to data types and use cases.

2. *See generally* Amy Winegar & Cass R. Sunstein, *How Much Is Data Privacy Worth? A Preliminary Investigation*, 42 J. CONSUM. POL'Y 425, 426 (2019); Xiau-Bai Li, Xiaoping Liu, X. & Louvai Motiwalla, *Valuing Personal Data with Privacy Consideration*, 52 DECISION SCI. 393, 417 (2021).

3. Big Tech usually refers to a group of the five largest technology companies in the world, primarily characterized by their substantial market cap and significant impact on the economy and society. Traditionally, the term is often used to describe the "Big Five" US tech companies: Alphabet (Google), Amazon, Apple, Meta (formerly Facebook), and Microsoft. Siddhesh Shinde, *What Companies Fall Under Big Tech? How Do You Land a Job With Them*?, EMERITUS (Nov. 14, 2024), https://emeritus.org/blog/technology-big-tech/ [https://perma.cc/3Z8R-N46V]. In the age of AI, it can be said that Big Tech also includes Nvidia and OpenAI. *See* Phil Rosen, *Big Tech Carries the Entire Stock Market — and Nvidia Fuels Big Tech*, INC. (Oct. 25, 2024), https://www.inc.com/phil-rosen/nvidia-stock-market-outlook-earnings-tech-magnificent-seven-economy-fed/90994387 [https://perma.cc/86RW-GPPU].

4. This dilemma has sometimes been described as "privacy paradox." *See* Alessandro Acquisti, Laura Brandimarte & George Loewenstein, *Secrets and Likes: The Drive for Privacy and The Difficulty of Achieving it in The Digital Age*, 30 J. CONSUMER PSYCH., 736, 749 (2020); Zohar Efroni, *Gaps and Opportunities: The Rudimentary Protection for "Data-Paying Consumers" Under New EU Consumer Protection Law*, 57 COMMON MKT. L. REV. 799, 799–802 (2020); Kirsten Martin, *Manipulation, Privacy, and Choice*, 23 N.C.J.L. & TECH. 452, 500–01 (2022); Axel Metzger & Heike Schweitzer, Shaping Markets: A Critical Evaluation of the Draft Data Act, 1 ZEUP (forthcoming 2023) (manuscript at 1) (on file with SSRN); Nina Gerber, Paul Gerber & Melanie Volkamer, *Explaining the Privacy Paradox: A Systematic Review of Literature Investigating Privacy Attitude and Behavior*, 77 COMPUT. & SEC. 226, 227 (2018).

situation apparently so different in the case of personal data?[5] Why is there such a disconnect between the proclaimed value data has for people and how they handle it?[6] Such questions require this Article to delve deeper into the intricate dynamics of data privacy and user behavior to address the disparity between people's perceived value of their data and their actions in handling it.[7] One possible explanation is the lack of practical tools that empower people with their own data.[8]

On a global scale, the emergence and proliferation of sensors, distributed cloud computing technologies, federated machine learning, and large language models place global society on a revolutionary trajectory in the ways in which we communicate with one another and interact with the technology that surrounds us.[9] Huge volumes of data powered by machine learning and large language models opened new opportunities to interact with data in multimodal dimensions, such as text to voice, text to image, voice to text, and more.[10] Imagine a world in which personal Artificial Intelligence (AI)-powered apps and AI agents operate on top of all the data a consumer has generated over the years. How will consumers utilize their consolidated historical data from various platforms and services through their personal AI-powered doctors, coaches, advisors, and intelligent agents?[11] These new, powerful AI-driven apps and intelligent agents that run on top of localized data sets are already being built and are becoming ubiquitous.[12] This Article explores what legal, technological, and ethical issues should be considered by policymakers and tech entrepreneurs developing these powerful and increasingly ubiquitous AI-powered applications.

---

5. *See* Igor Syunin, *Why are Consumers So Willing to Give up Their Data?*, DATAFLOQ (June 20, 2019), https://datafloq.com/read/why-are-consumers-willing-give-up-personal-data [https://perma.cc/2WHL-2YXC].

6. *See Decoding the Privacy Paradox and Balancing User Concerns with Online Behaviour*, ZEOTAP (Dec. 14, 2022), https://zeotap.com/blog/what-does-the-privacy-paradox-mean-for-the-online-industry/ [https://perma.cc/L87X-G6S2].

7. *See, e.g.*, Alessandro Acquisti, Laura Brandimarte & George Loewenstein, *Privacy and Behavioral Economics*, *in* MODERN SOCIO-TECHNICAL PERSPECTIVES ON PRIVACY 61–62 (Bart P. Knijnenburg et al. eds., 2022); *see also* Alessandro Acquisti, Laura Brandimarte & Jeff Hancock, *How Privacy's Past May Shape its Future*, 375 SCI. 270, 270–72 (2022); Sarah Turner & Leonie Maria Tanczer, *In Principle vs in Practice: User, Expert and Policymaker Attitudes Towards the Right to Data Portability in the Internet of Things*, 52 COMPUT. L. & SEC. REV. 1, 2 (2024).

8. David Cicilline, *Opening Statement at the House Judiciary Antitrust Subcommittee Hearing on Big Tech*, AM. RHETORIC (July 29, 2020), https://www.americanrhetoric.com/speeches/davidcicillineBIGTECHantitrusthearingopeningstatement.htm [https://perma.cc/R5RJ-U2V6] ("When everyday Americans learn how much of their data is being mined, they can't run away fast enough. But in many cases, there's no escape from this surveillance because there's no alternative. People are stuck without options."); Johann Kranz et al., *Data Portability*, 65 BUS. INFO. SYS. ENG'G. 597, 597 (2023), ("[U]sers are left with little meaningful options to adopt data protection and privacy measures and to move to rival OSPs due to the skewed playing field and high switching barriers.").

9. *See* Mark Fenwick & Paulius Jurcys, *Building a 'Green Data' Future: How a Human-Centric Approach to Data and Nudges can Help Fight Climate Change*, 18 J. INTELL. PROP. L. & PRAC. 386, 388–98, 394 (2023); *see also* GUIDO NOTO LA DIEGA, INTERNET OF THINGS AND THE LAW: LEGAL STRATEGIES FOR CONSUMER-CENTRIC SMART TECHNOLOGIES 1, 2, 92 (2022). For the implications of sensors and data to the medical services industry, *see Medtech and the Internet of Medical Things: How Connected Medical Devices are Transforming Health Care*, DELOITTE 1, 4 (July 2018), https://www2.deloitte.com/content/dam/Deloitte/global/Documents/Life-Sciences-Health-Care/gx-lshc-medtech-iomt-brochure.pdf [https://perma.cc/597L-LGJK]; H. Ceren Ates, Peter Q. Nguyen, Laura Gonzalez-Macia, Eden Morales-Narváez, Firat Güder, James J. Collins & Can Dincer, *End-to-End Design of Wearable Sensors*, 7 NATURE REV. MATERIALS 887, 887 (2022).

10. *See* Nikolaj Buhl, *Top 10 Multimodal Datasets*, ENCORD BLOG (Aug. 15, 2024), https://encord.com/blog/top-10-multimodal-datasets/ [https://perma.cc/BR62-TPMW].

There is an emerging consensus that data portability—and questions of data access and data interoperability, more generally—is an increasingly central issue in the new data economy.[13] As a result, regulatory frameworks are increasingly addressing these issues.[14] It is often argued that, at least from a policy perspective, data access and portability regimes can significantly promote innovation and facilitate competition.[15] Indeed, data portability is seen as one of the prerequisites for the development of new services and new markets.[16] At the same time, to maximize public welfare and protect individuals,

11. Paulius Jurcys, *The Personal AI Revolution: A Human-Centric Approach*, MEDIUM (July 31, 2023), https://medium.com/prifina/the-personal-ai-revolution-a-human-centric-approach-840f47e92a3a [https://perma.cc/BR62-TPMW]; *see, e.g.*, Mark Fenwick & Paulius Jurcys, *From Cyborgs to Quantified Selves: Augmenting Privacy Rights with User-Centric Technology and Design*, 13 J. INTELL. PROP., INFO. TECH & ELEC. COM. L. 20, 26 (2022).

12. *See, e.g.*, Fenwick & Jurcys, *supra* note 9, at 388; Paulius Jurcys, Ashley Greenwald, Mark Fenwick & Valto Loikkanen, Who Owns My AI Twin? Lights and Shadows of Data Ownership in a New World of Simulated Identities 3 (Sept. 30, 2024) (unpublished manuscript) (on file with SSRN), https://papers.ssrn.com/sol3/papers.cfm?abstract_id=4940663 [https://perma.cc/6JJW-TN5M].

13. Josef Drexl, *Connected Devices – An Unfair Competition Law Approach to Data Access Rights of Users*, *in* GERMAN FEDERAL MINISTRY OF JUSTICE AND CONSUMER PROTECTION MAX PLANCK INSTITUTE FOR INNOVATION AND COMPETITION, DATA ACCESS, CONSUMER INTERESTS AND PUBLIC WELFARE 477, 478 (2021).

14. *See id.*

15. Jiawei Zhang, *The Paradox of Data Portability and Lock-in Effects*, 36 HARV. J.L. & TECH. 658, 659 (2023); *Data Portability, Interoperability and Digital Platform Competition*, OECD (Apr. 27, 2022), http://oe.cd/dpic [https://perma.cc/2FSX-SJ3A]; Inge Graef, *The Opportunities and Limits of Data Portability for Stimulating Competition and Innovation*, CPI ANTITRUST CHRON. 1, 2 (2020); Orla Lynskey, *Aligning Data Protection Rights with Competition Law Remedies? The GDPR Right to Data Portability,* 42 EUR. L. REV. 793, 796 (2017); Alexandre de Streel, Jan Kraemer & Pierre Senellart, *Making Data Portability More Effective for the Digital Economy*, CTR. ON REGUL. EUR. L. 1, 9 (2020).

16. *See A European Strategy for Data*, EUR. COMM'N (Feb. 2, 2020), https://eur-lex.europa.eu/legal-content/EN/TXT/HTML/?uri=CELEX:52020DC0066 [https://perma.cc/76DW-7E6N].

it is necessary to effectively balance the diverse interests of consumers, businesses, and other stakeholders operating in specific industries.[17] However, most of the academic research on data portability is focused on Article 20 of the General Data Protection Regulation (GDPR) and subsequent European Union regulations.[18] This Article advances the discussion by offering a broader interdisciplinary perspective and focusing on the economic incentives and emerging technological models that underpin the implementation of data portability rights.

More specifically, this Article critically examines the current regulatory discourse surrounding data portability, taking recent developments in the European Union as an example. While regulation is welcome, the current implementation primarily reinforces the interests of service providers, data controllers, and data enterprises rather than prioritizing the rights and interests of end users. To restore public trust, a paradigm shift towards a more human-centric data ecosystem is vital, where data is emancipated from service-provider-controlled silos and placed in the hands of individuals. In a human-centric data ecosystem, in a world where data is central to our identities and the kind of subjects we are, data needs liberating.[19]

17. Inge Graef, Martin Husovec & Nadezhda Purtova*, Data Portability and Data Control: Lessons for an Emerging Concept in EU Law*, 19 GER. L. J., 1359, 1398 (2018); *see* Metzger & Schweitzer, *supra* note 4, at 11; *see, e.g.*, Simonetta Vezzoso, *Competition Policy in Transition: Exploring Data Portability's Roles*, 12(4) J.E.C.L. & PRACT. 368–69 (2021).

18. *See, e.g.*, Paul De Hert, Vagelis Papakonstantinou, Gianclaudio M., L. Beslay & I. E. Sanchez, *The Right to Data Portability in the GDPR: Towards User-Centric Interoperability of Digital Services*, 34 COMPUT. L. & SEC. REV. 193, 196–202 (2018); Helena U. Vrabec, *Data Portability as a Data Subject Right*, *in* DATA SUBJECT RIGHTS UNDER THE GDPR 159, 159–60 (2021); Wenlong Li, *A Tale of Two Rights: Exploring the Potential Conflict Between Right to Data Portability and Right to be Forgotten Under the General Data Protection Regulation*, 8 INT'L. DATA PRIV. L. 309, 309–11 (2018). For empirical studies exploring practical challenges related to the exercise of data access and portability rights *see, e.g.*, Janis Wong & Tristan Henderson, *The Right to Data Portability in Practice: Exploring the Implications of the Technologically Neutral GDPR*, 9 INT'L. DATA PRIV. L. 173, 174 (2019); Emmanuel Syrmoudis, Stefan Mager, Sophie Kuebler-Wachendorff, Paul Pizzinini, Jens Grossklags & Johann Kranz, *Data Portability Between Online Services: An Empirical Analysis on the Effectiveness of GDPR Art. 20*, 3 PROC. PRIV. ENHANCING. TECHS. 351, 352 (2021); Sophie Kuebler-Wachendorff, Robert Luzsa, Johann Kranz1, Stefan Mager1, Emmanuel Syrmoudis, Susanne Mayr & Jens Grossklags, *The Right to Data Portability: Conception, Status Quo, and Future Directions*, 44 INFORMATIK SPEKTRUM 264, 266 (2021); *see generally* Matthias Leistner & Lucie Antoine, *IPR and the Use of Open Data and Data Sharing Initiatives by Public and Private Actors,* EUR. PARLIAMENT 1, 34 (May 3, 2022), https://www.europarl.europa.eu/thinktank/en/document/IPOL_STU(2022)732266 [https://perma.cc/53SM-ECLJ].

19. *See* Mark Fenwick, Paulius Jurcys & Aidas Liaudanskas, Voice Cloning in an Age of Generative AI: Mapping the Limits of the Law & Principles for a New Social Contract with Technology 1, 6 (Aug. 24, 2024) (unpublished article) (on file with SSRN), https://papers.ssrn.com/sol3/papers.cfm?abstract_id=4850866 [https://perma.cc/U7EV-HYYS]; Jurcys et al., *supra* note 12, at 26.

In short, concepts of data portability must be reimagined as fundamental freedoms and not super qualified rights that offer limited control over information about us.

Part II introduces how data access and portability rights are framed in current regulations such as the EU's GDPR, Data Act, Digital Markets Act (DMA), Data Spaces Regulation, and AI Act. These laws give individuals the right to request that companies provide access to their data and "port" this data from one service provider ("Service Provider A") to another ("Service Provider B").[20] The rights to access and port one's personal data are typically presented as empowering individuals by offering a new degree of control over personal data.[21] Introducing data portability is the foundation—or at least a trigger—for a shift from today's enterprise-centric, siloed, and sealed data ecosystem to a more open, human-centric vision of the future.[22]

In Part III, this Article reveals that current notions of data access and portability—as articulated in a European context—frame our relationship with data in an overly narrow or restrictive way. The European Union represents data as something that exists separately from persons—data *about* persons or the subjects *of* data.[23] The data then moves around the siloed, enterprise-controlled, and product-centric data ecosystem under the partial (and often illusory) control of individuals—control that is ultimately negotiated with data controllers, which are typically third-party corporations.[24] This type of relationship is overly restrictive and sets portability expectations too low.[25]

Instead, taking inspiration from technological trends already occurring in the European Union and elsewhere,[26] this Article proposes something more legally and technically radical than the legal concept of data portability enshrined in GDPR and California Consumer

20. *See* Regulation 2016/679, of the European Parliament and of the Council of 27 April 2016 on the Protection of Natural Persons with regard to the Processing of Personal Data and on the Free Movement of Such Data and Repealing Directive 96/46/EC (General Data Protection Regulation), 2016 O.J. (L 119) 1, 43, 45 (EU) [hereinafter General Data Protection Regulation].

21. *See, e.g.*, *id.* at 2 (emphasizing the "importance of creating the trust that will allow the digital economy to develop across the internal market" and positing that "natural persons should have control of their own personal data.").

22. *See* Sille Sepp, *A Human-Centric Approach to Personal Data*, MEDIUM (Oct. 2, 2023), https://medium.com/opendatacharter/a-human-centric-approach-to-personal-data-b73268474851 [https://perma.cc/9FPS-XLHH].

23. *See* General Data Protection Regulation, *supra* note 20, at 33.

24. *See* Daniel Solove, *Murky Consent: An Approach to the Fictions of Consent in Privacy Law*, 104 B.U.L. REV. 593, 622 (2024).

25. *See id.* at 604–05.

26. *See* discussion *infra* Part III.

Privacy Act (CCPA): specifically, the freedom to gather data together in a secure personal space under individual users' dominion and control.[27] This Article develops this framework by describing the type of technical architecture necessary to facilitate this more ambitious version of portability through three different technical models of data portability: (i) siloed (consent and control), (ii) intermediary (enterprise-centric), and (iii) human-centric frameworks. Each model is progressively more empowering for the individual user. However, a shift toward the third, human-centric data model is not only the most desirable but also more technologically feasible than ever before.[28] The third model can address public concerns around personal data and diminishing trust in technology and technology firms by offering individuals dominion over "their" data.[29]

Finally, Part IV concludes with some reflections on the meaning of portability and how a more expansive concept of this term can provide impetus for the technological trends introduced in Part III.

## II. REGULATORY CONCEPTS & TRENDS IN DATA PORTABILITY: EUROPEAN EXAMPLES

This section outlines the main features of current and emerging legal frameworks establishing rights of data access and portability with a focus on selected stipulations in the evolving regulatory landscape of the European Union. Due to its long-standing history and sophistication, both in terms of regulations and related debates, the European Union has been a global "forerunner" in many areas of data protection and AI regulation.[30] Given the geopolitical context and emerging international competition to develop state-of-the-art data and AI technologies, the European Union aims to provide a legal framework for data markets to open novel opportunities for data-driven

27. California Consumer Privacy Act of 2018, CAL. CIV. CODE §§ 1798.100–1798.199.100; W. Gregory Voss, *The CCPA and the GDPR Are Not the Same: Why You Should Understand Both*, 1 CPI ANTITRUST CHRON. 1, 7–12 (2021).

28. *See* discussion *infra* Section III C.

29. *See id.*

30. *See, e.g.*, Michael L. Rustad and Thomas H. Koenig, *Towards a Global Data Privacy Standard*, 71 FLA. L. REV. 365, 370–75, 389 (2019) (exploring the "Brussels Effect" and also highlighting the overlooked "D.C. Effect"); René Mahieu, Hadi Asghari, Christopher Parsons, Joris van Hoboken, Masashi Crete-Nishihata, Andrew Hilts & Siena Anstis, *Measuring the Brussels Effect Through Access Requests: Has the European General Data Protection Regulation Influenced the Data Protection Rights of Canadian Citizens?*, 11 J. INFO. POL'Y 301, 335 (2021); *see also* Shannon Williams, *EU AI Act Sets New Global Standard for Ethical AI Use*, ITBRIEF (July 31, 2024), https://itbrief.co.uk/story/eu-ai-act-sets-new-global-standard-for-ethical-ai-use [https://perma.cc/N3LN-2TGJ].

competition and innovation proactively.[31] The existing and upcoming EU legislation is particularly well suited to illustrate the main features of evolving data access and portability rights.

Recently, a particularly intricate and complex regulatory ecosystem has rapidly evolved in the European Union.[32] The current regulatory landscape is challenging due to many interacting and overlapping regulations that are already enacted or proposed and will soon be adopted after additional negotiations and potential changes.[33] However, the most relevant for this Article are the GDPR, the Data Act, the DMA, the European Health Data Space (EHDS) Regulation, and the AI Act.[34] This section highlights the key tenets of these developments and how they pave the way for the emergence of new approaches and solutions to implement data access and portability regimes in practice.

### *A. Data Access and Portability Under the GDPR*

The GDPR is a far-reaching and comprehensive data protection framework that came into force in May 2018.[35] The GDPR's main goal is to protect the privacy and data rights of individuals within the European Union, but its territorial scope also extends to the use of European data elsewhere.[36] More specifically, the GDPR includes stipulations that grant data access rights and portability rights to strengthen individuals' control over their personal data and safeguard transparency requirements in data processing activities within and outside of the European Union.[37]

---

31. *See Communication from the Commission to the European Parliament, the Council, the European Economic and Social Committee and the Committee of the Regions, A European Strategy for Data*, EUR. COMM'N (Feb. 19, 2020), https://eur-lex.europa.eu/legal-content/EN/TXT/HTML/?uri=CELEX:52020DC0066 [https://perma.cc/868S-W5C8]*; see also* Florent Thouvenin & Aurelia Tamò-Larrieux, *Data Ownership and Data Access Rights: Meaningful Tools for Promoting the European Digital Single Market?*, *in* BIG DATA AND GLOBAL TRADE LAW 316, 319, 331 (Mira Buri ed., 2021).

32. For an in-depth overview, *see* HEIKE SCHWEITZER, AXEL METZGER, KNUT BLIND, HEIKO RICHTER, CRISPIN NIEBEL & FREDERIK GUTMANN, DATA ACCESS AND SHARING IN GERMANY AND IN THE EU: TOWARDS A COHERENT LEGAL FRAMEWORK FOR THE EMERGING DATA ECONOMY, 1, 44–51 (2022).

33. *See, e.g.*, Shannon Yavorski & Jeremy Kudon, *The Future of AI Regulation and Legislation: 5 Key Takeaways*, ORRICK (Sept. 23, 2024), https://www.orrick.com/en/Insights/2024/09/The-Future-of-AI-Regulation-and-Legislation-5-Key-Takeaways [https://perma.cc/Y3S6-PRWB] (hosting an expert panel discussion on the potential impact of more than seven hundred currently pending AI-related legislative proposals).

34. For more sector-specific ramifications of data portability *see, e.g.*, Daniel Gill & Wolfgang Kerber, Data Portability Rights: Limits, Opportunities, and the Need for Going Beyond the Portability of Personal Data 1, (Oct. 11, 2020), (unpublished article) (on file with SSRN). *See generally* Can Atik, *Towards Comprehensive European Agricultural Data Governance: Moving Beyond the "Data Ownership" Debate*, 53 INT'L. REV. INTELL. PROP. & COMPETITION L. 701 (2022).

35. *See* General Data Protection Regulation, *supra* note 20, at 87.

36. *See, e.g.*, *id.* at 32 (defining the territorial scope of application of GDPR).

### 1. Data Access Rights Under the GDPR

The rights enshrined in Article 15 of the GDPR allow individuals to understand how their data is used and ensure the lawfulness and fairness of such processing activities. It grants every data subject the right to "obtain from the controller confirmation as to whether or not personal data concerning him or her are being processed."[38] Article 15 entitles individuals to request access to more specific information, such as the purposes of the processing, the categories of personal data involved, the recipients or categories of recipients to whom the data has been or will be disclosed, and the envisaged retention period.[39] Data controllers must respond to such access requests within one month,[40] and any unjustified refusal or delay may result in penalties.[41]

To facilitate the process and support the data subject's exercise of rights, data controllers must also provide a copy of the personal data undergoing processing to the individual, usually free of charge.[42] These controller-provided records should enable individuals to easily understand and analyze the data being processed in a commonly used and machine-readable format.[43] For "any further copies requested by

---

37. *See id.* at 43, 45 (defining data access rights and data portability rights).

38. *Id.* at 33, 43 (explaining that the definition of "personal data" in the GDPR is broad and includes "any information relating to an identified or identifiable natural person.").

39. *Id.* at 43.

40. *Id.* at 40.

41. *Cf. id.*

42. *Id.* at 40, 43. The scope of data access under the GDPR covers only input and metadata and does not cover observed or observable and derived data. For a discussion, *see, e.g.*, Jan Krämer, *Personal Data Portability in the Platform Economy: Economic Implications and Policy Recommendations*, 17 J. COMPETITION L. & ECON. 1, 2 (2021); *cf.* General Data Protection Regulation, *supra* note 20, at 40. Article 12(5) stipulates that "[w]here requests from a data subject are manifestly unfounded or excessive, in particular because of their repetitive character, the controller may either: (a) charge a reasonable fee taking into account the administrative costs of providing the information or communication or taking the action requested; or (b) refuse to act on the request." General Data Protection Regulation, *supra* note 20. at 40.

43. General Data Protection Regulation, *supra* note 20, at 43; *cf. id.* at 39–40. However, since most data access requests are provided in machine-readable JSON format, the use of such data is of limited practical value to an average consumer. *See Digital PDF vs. Machine-Readable JSON Format*, PLANET AI (Nov. 28, 2024) https://planet-ai.com/digital-pdf-vs-machine-readable-json-format/#:~:text=the%20data%20inside.-,JSON%20%E2%80%93%20The%20Machine%2Dreadable%20Format%20for%20Structured%20Data,be%20easily%20processed%20by%20machines. [https://perma.cc/G4X2-7BLY].

the data subject, the controller may charge a reasonable fee based on administrative costs."[44]

2. Data Portability Under the GDPR

Where individuals have provided personal data to the data controller based on consent or for the performance of a contract,[45] Article 20, Section 1 of the GDPR grants individuals the right to receive their personal data in a "structured, commonly used, and machine-readable format."[46] Individual consumers also have the right to "transmit those data to another data controller without hindrance from the data controller to which the personal data have been provided."[47] Moreover, by obliging data controllers to transfer the personal data directly to another controller when technically feasible, individuals are enabled to exercise their right to portability without undue and unnecessary obstacles.[48] In other words, Article 20 in combination with related provisions in Articles 12–18, provides individuals with the opportunity to choose and switch among different services and platforms while minimizing barriers to entry.[49] The ability to choose and switch enables seamless transfers and allows data subjects to remain in control of their personal data and the way it is used while also facilitating competition among service providers.[50]

There is one caveat that highlights the need to strengthen the data access, interoperability, and data transfer systems to promote a more effective exercise of individual rights under the GDPR. In the absence of any special rule (*lex specialis*), data controllers are generally not required to adopt or maintain data processing and transfer systems that are technically compatible with other controllers in different organizations.[51] The existing regulatory framework under the GDPR seems to reveal the current insufficiency of so-called existing data portability mechanisms. Technology incumbents might point to the status quo as "Portability Possibility."[52] However, a more accurate description of the situation could be seen as "Portability Impossibility," which cannot be remedied unless there is "Portability Compatibility."[53]

---

44. General Data Protection Regulation, *supra* note 20, at 43.

45. *Id.* at 45 (explaining that where "the processing is based on consent pursuant to point (a) of Article 6(1) or point (a) of Article 9(2) or on a contract pursuant to point (b) of Article 6(1); and the processing is carried out by automated means.").

46. *Id.*

47. *Id.*

48. *See id*.

49. *See* Daniel Rubinfeld, *Data Portability and Interoperability: An E.U.-U.S. Comparison*, 57 EUR. J.L. ECON. 163, 164 (2024).

50. *See id.* at 173.

51. *See* General Data Protection Regulation, *supra* note 20, at 45.

### *B. The EU Data Act*

In an evolving European legal framework concerning data protection and access, the European Data Act,[54] which complements the Data Governance Regulation,[55] stands as one of the most remarkable legislative achievements for being the first comprehensive regulation of such broad scope.[56] The Data Act sets its sights on bolstering the European Union's data economy by "unlocking industrial data, optimizing its accessibility and use, and fostering a competitive and reliable European cloud market,"[57] while also "ensur[ing] that the benefits of the digital revolution are shared by everyone."[58] The Data Act aims to remove possibilities for lock-in of data generated by Internet of Things (IoT) devices, to enhance data accessibility and utility, and to stimulate a competitive and trustworthy data economy in Europe.[59]

The Data Act introduces a new set of data access rights for IoT product users and the third parties authorized by them.[60] IoT device users have the right to access data generated by their use of the IoT product.[61] One of the Act's most remarkable features is that the concept

---

52. *See id.*

53. *See id.*

54. Regulation (EU) 2023/2854 of the European Parliament and of the Council of 13 December 2023 on Harmonized Rules on Fair Access to and Use of Data and Amending Regulation (EU) 2017/2394 and Directive (EU) 2020/1828 (Data Act), 2023 O.J. (L 119) 1–71 [hereinafter Data Act].

55. Regulation (EU) 2022/868 of the European Parliament and of the Council of 30 May 2022 on European Data Governance and Amending Regulation (EU) 2018/1724 (Data Governance Act), 2022 O.J. (L 152) 1–44 [hereinafter Data Governance Act].

56. For a more detailed analysis, *see* Metzger & Schweitzer, *supra* note 4 at 2; Charlotte Ducuing, Thomas Margoni & Luca Schirru, *White Paper on the Data Act Proposal* 10 (KU Leuven Centre for IT & IP Law, Working Paper, 2022), https://www.law.kuleuven.be/citip/en/Publications/citip-whitepaperdataact.pdf [https://perma.cc/TA49-ZZE6]; Rupprecht Podszun & Philipp Offergeld, The EU Data Act and the Access to Secondary Markets 9 (unpublished manuscript) (Nov. 1, 2022) (on file with SSRN); Leistner & Antoine, *supra* note 18, at 71.

57. European Commission Press Release IP/23/3491, Data Act: Commission Welcomes Political Agreement on Rules for a Fair and Innovative Data Economy (June 27, 2023).

58. *Id.*

59. *See* Data Act, *supra* note 54, at 11. For a more critical analysis, *see generally, e.g.*, Wolfgang Kerber, *Governance of IoT Data: Why the EU Data Act Will not Fulfil Its Objectives*, 72 GRUR INT. 1, 1–16 (2022).

60. *See* Data Act, *supra* note 54, at 37.

61. *Id.* at 37 ("Connected products shall be designed and manufactured, and related services shall be designed and provided, in such a manner that product data and related service data, including the relevant metadata necessary to interpret and use those data, are, by default, of "data users" is defined broadly: it encompasses individual consumers and business entities.[62] Article 4(1) of the Act places data holders under an obligation to grant the user of the device access to the data generated by the use of a product or related service.[63] Such access to data should be provided without undue delay, free of charge, and where applicable, continuously and in real time.[64] Moreover, under Article 5, the data holder shall make such data available not only to the user of the device but also to a party acting on behalf of a user.[65] In situations where a third party is acting on behalf of the user, the processing of such data is subject to "the conditions agreed with the user."[66]

The material scope of the Data Act is limited to data generated by IoT devices and services related to such devices.[67] Notably, those data access rights apply vis-à-vis data holders regardless of whether they are dominant "gatekeepers" in the market.[68] Furthermore, the scope of data accessible under the Data Act covers only individual-level data and does not cover bundled or aggregated individual-level data.[69] By doing so, the Data Act aims to establish a private law infrastructure of "horizontal" legal rights over individual nonexclusive rights of data access.[70] Such a right exists "by default" regardless of the underlying easily, securely, free of charge, in a comprehensive, structured, commonly used and machine-readable format, and, where relevant and technically feasible, directly accessible to the user.").

62. *Id.* at 34.

63. *Id.* at 38 ("[W]here data cannot be directly accessed by the user from the connected product or related service, data holders shall make readily available data, as well as the relevant metadata necessary to interpret and use those data, accessible to the user without undue delay, of the same quality as is available to the data holder, easily, securely, free of charge, in a comprehensive, structured, commonly used and machine-readable format and, where relevant and technically feasible, continuously and in real-time.").

64. *Id.*

65. Data Act, *supra* note 54, at 40.

66. *Id.*, at 41.

67. *Id.*, at 4. ("[C]onnected products that obtain, generate or collect, by means of their components or operating systems, data concerning their performance, use or environment and that are able to communicate those data via an electronic communications service, a physical connection, or on-device access, often referred to as the Internet of Things, should fall within the scope of this Regulation.").

68. Heike Schweitzer & Axel Metzger, *Data Access under the Draft Data Act, Competition Law and the DMA: Opening the Data Treasures for Competition and Innovation?*, 72 GRUR INTL. 337, 355 (2003); see also Data Act, *supra* note 54, at 32–33.

69. Schweitzer & Metzger, *supra* note 68, at 344.

70. *Id.* at 340. In the context of the Data Act, "horizontal" rights to access data refer to a broad and non-sector-specific right that allows both private and business users to access data generated by the use of products or related services. *See* Data Act, *supra* note 54, recital 42, at 14. This approach aims to establish a uniform framework for data access that is not limited to specific sectors, promoting fairness and innovation across various industries. *Id.*

contractual arrangement between the product user and the data holder.[71]

### *C. The EU Digital Markets Act*

The DMA came into force on November 1, 2022.[72] It aims to make digital markets safer, fairer, more open, and contestable.[73] The DMA aims to alleviate (1) existing concerns that digital markets are not functioning properly and (2) the economic harms that stem from the fact that the digital space is dominated by "gatekeeper" platforms.[74] Accordingly, the DMA applies to the activities of the gatekeepers, a term which covers core platform services such as online search engines, social networking services, video-sharing platforms, and web browsers.[75] The additional constraints found within the DMA are meant to supplement already-existing obligations under European competition law,[76] to remedy economic imbalances, unfair business practices, and their negative impact on the contestability of platform markets.[77]

The European Commission is entrusted to designate which companies fall under the definition of gatekeeper and must comply with obligations established in the DMA.[78] For instance, Article 5 of the DMA prohibits combining personal data from the core service with third-party services.[79] Gatekeepers are also prohibited from cross-using personal data between the core service and other services.[80] However, these restrictions would not apply if individuals give their consent to the combination and cross-use of data.[81]

---

71. Data Act, *supra* note 54, at 37.

72. Regulation (EU) 2022/1925 of the European Parliament and of the Council of 14 September 2022 on Contestable and Fair Markets in the Digital Sector (Digital Markets Act), art. 54, 2022 O.J. (L 265) 1–66 [hereinafter Digital Markets Act].

73. *The Digital Services Act Package*, EUR. COMM'N (Oct. 4, 2024), https://digital-strategy.ec.europa.eu/en/policies/digital-services-act-package [https://perma.cc/NU2F-NQR2].

74. Digital Markets Act, *supra* note 72, recitals 2–11; *see also* Björn Lundqvist, *The Proposed Digital Markets Act and Access to Data: A Revolution, or Not?*, 52 INT'L REV. INTELL. PROP. & COMPETITION L. 239, 239–41 (2021).

75. *See* Digital Markets Act, *supra* note 72, art. 2(1) at 28. The definition of a "gatekeeper" in the EU DMA is rather complex; "gatekeepers" are defined as large online platforms that have a significant impact on the internal market, serve as an important gateway for businesses to reach consumers, and possess a strong economic position. Specifically, gatekeepers are typically designated based on criteria such as their size, the number of users, and their role in providing core platform services. The DMA aims to ensure that these gatekeepers do not abuse their market power and that they provide fair access to their platforms for other businesses. *See id.* at 30–32.

76. *See, e.g.*, Digital Markets Act, *supra* note 72, at 1–3.

77. *See id.* at recital 6 at 2.

78. *Id.* at 31.

79. *Id.* at 33

80. *Id.*

Under the DMA, gatekeepers must facilitate, free of charge, the effective and high-quality portability of data that is provided by the end user or generated through the activity of the "end user in the context of the use" of the platform.[82] Such data portability rights are provided for individual consumers and business users of the platform.[83] The data should be made available in a format that "can be immediately and effectively assessed and used by the user or the relevant third party authorized by the end user to which the data is ported."[84]

Accordingly, gatekeepers must provide data portability tools such as Application Programming Interfaces (APIs) that allow continuous and real-time access to data.[85] The gatekeeper must provide access to personal data only where it is directly connected with the use effectuated by the end users with respect to the products or services offered by the relevant business user through the relevant core platform service, and when the end users provide consent.[86] Failure to comply with the obligations may result in fines up to 10 percent of the gatekeeper's worldwide turnover.[87] This amount refers to the total amount of revenue generated by a company.

### *D. The Proposed EU Health Data Space Regulation*

Since the adoption of the GDPR, the European Commission has discussed the legislative objectives outlined in the Data Strategy and the recent Data Governance Act.[88] The Commission aims to support the creation of personal data spaces in different sectors such as agriculture, health, and mobility.[89] Such sector-specific data spaces are based on the common idea of strengthening the right to portability enshrined in the GDPR.[90] One of these proposed sector-specific regulations applies in combination with and based on other fundamental regulations, such as the GDPR or the Medical Device Regulation (MDR),[91] but specifically targets special data spaces.[92] The recently proposed EHDS regulation,[93] for example, seeks to establish a framework for the secure and interoperable exchange of health data across the European Union.[94] This includes specific access and data portability rights for individuals.[95]

---

81. *Id.*
82. *Id.* at 36.
83. *Id.*
84. *Id.* at 15.
85. *Id.* at 36.
86. *Id.*
87. *Id.* at 51.
88. For an initial overview, *see generally European Health Data Space*, EUR. COMM'N, https://health.ec.europa.eu/ehealth-digital-health-and-care/european-health-data-space_en [https://perma.cc/Y9UJ-SAZE] (last visited Feb. 11, 2025).
89. *See Communication from the Commission to the European Parliament*, *supra* note 31, at 6.
90. *Id.* at 10.

#### 1. Data Access Under the Proposed EHDS Regulation

One of the goals of the proposed EHDS regulation is to promote the ability of individuals to access their primary health data kept by health authorities, healthcare providers, or other relevant entities.[96] Article 3 of the proposed EHDS regulation stipulates that natural persons "shall have the right to access their personal electronic health data processed in the context of primary use of electronic health data, immediately, free of charge and in an easily readable, consolidated and accessible form."[97] This also implies that natural persons can request and receive copies of their health data, such as medical records, test results, and treatment information.[98]

To safeguard and support effective data access, Article 3 and further stipulations in the proposed EHDS regulation set frameworks, requirements, and standards to push the use of digital health tools and electronic health records.[99] These tools and frameworks mandated by the EHDS regulation are supposed to make it easier for natural persons to access their health data online, by logging into secure portals or mobile applications.[100] By promoting mechanisms and requiring the use of electronic health records and new technologies to provide easier and more expedient access to health data, the EHDS regulation would enable natural persons to better and more actively participate in monitoring their health statuses, reaching their own healthcare decisions, and sharing relevant data with healthcare professionals of their choice.[101]

---

91. Commission Regulation 2017/745 on Medical Devices, Amending Directive 2001/83/EC, Regulation (EC) No 178/2002 and Regulation (EC) No. 1223/2009 and Repealing Council Directives 90/385/EEC and 93/42/EEC, 2017 O.J. (L 117) 1–175.

92. *See Common European Data Spaces*, EUR. COMM'N (Jan. 23, 2025), https://digital-strategy.ec.europa.eu/en/policies/data-spaces [https://perma.cc/9MK3-4JUG].

93. *Proposal for a Regulation of the European Parliament and of the Council on the European Health Data Space (EHDS)*, COM (2022) 197 final (May 3, 2022) [hereinafter Proposed EHDS Regulation].

94. *See id.* at 4 (Explanatory Memorandum).

95. *Id.* at 24–25. For an overview of various data portability solutions in the healthcare spaces, *see* René Raab, Arne Küderle, Anastasiya Zakreuskaya, Ariel D Stern, Jochen Klucken, Georgios Kaissis, Daniel Rueckert, Susanne Boll, Roland Eils, Harald Wagener & Bjoern M. Eskofier, *Federated Electronic Health Records for the European Health Data Space*, 5 LANCET DIGIT. HEALTH e840, e841 (2023); Andreas Panagopoulos, Timo Minssen, Katerina Sideri, Helen Yu & Marcelo Corrales Compagnucci, *Incentivizing the Sharing of Healthcare Data in the AI Era*, 45 COMPUT. L. & SEC. REV. 1, 2 (2022).

96. Proposed EHDS Regulation, *supra* note 93, at 21.

97. *Id.* at 48.

98. *See id.*

99. *See id.* at 48–50; *see also, e.g.*, *id.* at 50–51.

Moreover, the proposed EHDS regulation contains stipulations that would regulate access by health professionals to personal electronic health data[102] and secondary use health data[103] under the governance of health data access bodies.[104]

2. Data Portability Under the Proposed EHDS Regulation

To improve the progression of care and enable individuals to freely select their healthcare providers, the EHDS regulation further acknowledges the increasing significance of data portability.[105] In the EHDS context, this refers to the right of natural persons to transfer their primary health data from one healthcare provider to another, or to other authorized entities. Moreover, the EHDS aims to improve GDPR-compliant data flows and data interoperability among scientists, healthcare providers, and authorities, which includes secondary uses (e.g., using health data for research purposes).[106] In providing a framework for the secondary use of electronic health data, the EHDS builds upon the proposed Data Governance Act and the proposed Data Act discussed above.[107]

The proposed EHDS regulation would also provide for the use of standardized formats and interoperable systems for promoting health data exchange and data portability.[108] To facilitate the seamless integration and transfer across different healthcare systems, the regulation would stipulate that health data should be coded and structured in a consistent and coherent manner.[109] In addition to facilitating health research and more effective healthcare systems, the

100. *See id.* at 27, 49.
101. Proposed EHDS Regulation, *supra* note 93, at 22, 48.
102. *Id.* at 50.
103. *Id.* at 68–71.
104. *Id.* at 71; *see also id.* at 75.
105. *Id.* at 24–25.
106. *See id.* at 24–25, 68–69.
107. *Id.* at 4 (Explanatory Memorandum).
108. *E.g.*, *id.* at 31, 38, 51.
109. *Id.* at 27, 48, 50–51.

proposed EHDS regulation also aims to facilitate better coordination and continuity of care by empowering individuals and patients to more easily switch among healthcare providers without losing access to their health data.[110]

Furthermore, the EHDS regulation would introduce the concept of the "MyHealth@EU" platform and call for its implementation, which would ultimately be mandatory in all Member States.[111] MyHealth@EU is a "cross-border infrastructure for primary use of electronic health data formed by the combination of national contact points for digital health and the central platform for digital health."[112] MyHealth@EU is supposed to provide a personalized and secure space, where natural persons can store and manage their health data.[113] Natural persons would be able to effectively share their personal electronic health data in the language of the country of destination when travelling abroad or take their personal electronic health data with them when moving to another country.[114] They would have control over their data space, deciding who can access their health data and for what purposes.[115] In addition, the EHDS calls for a similar infrastructure for secondary use of health data called "HealthData@EU." HealthData@EU is an "infrastructure connecting national contact points for secondary use of electronic health data and the central platform."[116]

Unsurprisingly, the proposed EHDS regulation continuously stresses the significance of privacy and data protection in the context of data access and portability throughout the entire document. Any processing of health data would have to comply with the GDPR and the requirements of any further relevant data protection laws.[117] This would also imply the need for implementing appropriate security measures to prevent unauthorized access or disclosure and the consent of data subjects for the processing of their personal health data.[118] The proposed EHDS regulation represents EU policymakers' ambitious vision to enhance data accessibility and portability by creating a common framework that would allow citizens to access and share their health data seamlessly across borders, while also facilitating the secure

110. *See id.* at 25, 48.
111. *Id.* at 29, 55–56; *see also id.* at 9 (Explanatory Memorandum).
112. *Id.* at 47.
113. *See id.* at 29.
114. *Id.* at 29; *see also id.* at 15 (Explanatory Memorandum).
115. *Id.* at 49, 55.
116. *Id.* at 47.
117. *Id.* recital 6 at 22–23; *see also id.* at 15 (Explanatory Memorandum).
118. *Id.* at 82–83.

use of health data for research, innovation, and policymaking throughout the European Union.

### *E. The EU AI Act*

Finally, the European Union is vigorously working on developing regulations to govern the use of AI technologies. The current discussions and proposals regarding a directly applicable EU AI regulation have been characterized by long-standing debates over the promises, perils, and trade-offs of underregulating and overregulating technologies, the introduction of new risk-category-based frameworks establishing liability rules and AI governance considerations (such as regulatory sandboxes), and inherent difficulties to cope with the rapid evolution of technologies.[119]

The AI Act, proposed in April 2021, was adopted on May 21, 2024.[120] It seeks to regulate the use of AI systems within the European Union and aims to promote the development of a human-centric approach to AI.[121] In June 2023, the AI Act proposal was amended by the European Parliament to incorporate rapidly evolving technological developments, such as in generative AI and large language models.[122] While the AI Act primarily focuses on AI-specific provisions, it builds upon and refers to several GDPR stipulations and related regulations such as the proposed EHDS and the enacted MDR.[123] Accordingly, the AI Act also affirms and strengthens data access and data portability rights to safeguard individuals' control over their personal data regarding AI-driven technologies, applications, and settings.[124] For example, Recital 165 of the AI Act stresses that "the Commission may develop initiatives, including of a sectoral nature, to facilitate the lowering of technical barriers hindering cross-border exchange of data for AI development, including on data access infrastructure, semantic and technical interoperability of different types of data."[125]

---

119. *See, e.g.*, Philip Hacker, Sustainable AI Regulation 1, 22 (Dec. 21, 2023) (unpublished manuscript) (on file with SSRN); Johann Laux, Sandra Wachter & Brent Mittelstadt, Trustworthy Artificial Intelligence and the European Union AI Act: On the Conflation of Trustworthiness and the Acceptability of Risk 6 (Sept. 26, 2022) (unpublished article) (on file with SSRN), https://papers.ssrn.com/sol3/papers.cfm?abstract_id=4230294 [https://perma.cc/9M8E-ZGV3]; Martin Ebers, Truly Risk-Based Regulation of Artificial Intelligence – How to Implement the EU's AI Act 12, 15 (June 19, 2024) (unpublished article) (on file with SSRN), https://papers.ssrn.com/sol3/papers.cfm?abstract_id=4870387 [https://perma.cc/4S9H-2RA6].

120. *Historic Timeline*, EU ARTIFICIAL INTEL. ACT, https://artificialintelligenceact.eu/developments/#:~:text=12%20July%202024%20%E2%80%93%20The%20AI,adopted%20the%20EU%20AI%20Act [https://perma.cc/M8WX-NU9R] (last visited Jan. 20, 2025).

121. Council Regulation 2024/1689 art. 1(1), 2023 O.J. (L 1689), 44 (EU) [hereinafter EU AI Act].

122. *See* Artificial Intelligence Act, EUR. PARL. DOC. P9_TA 0236 (2023), amend. 19, recital 6(a) [hereinafter Artificial Intelligence Act Proposal].

123. *See* EU AI Act, *supra* note 121, at 3–4, 12–13.

124. *See id.*

The revised AI Act proposal of June 14, 2023 is particularly interesting because it introduced the human-centric approach to data and AI.[126] Such a human-centric approach is the foundational cornerstone of how the European Union now regulates AI systems.[127] Recital 4(a) of the proposal provided that: "As a prerequisite, AI should be a human-centric technology. It should not substitute human autonomy or assume the loss of individual freedom and should primarily serve the needs of the society and the common good."[128] According to the EU legislature, such a human-centric approach to AI is rooted in the values of the Charter of Fundamental Rights of the European Union and the values upon which the Union is founded.[129] These values include the protection of fundamental rights, human agency and oversight, technical robustness and safety, privacy and data governance, transparency, non-discrimination and fairness, and societal and environmental wellbeing.[130]

Although the proposed provision in Recital 4(a) has not survived the legislative process, Article 1 of the adopted AI Act mandates all operators to make their best efforts to develop and use AI systems that promote a coherent human-centric European approach to ethical and trustworthy AI.[131] The EU regulation is technologically agnostic: it does not provide any clear technological solutions concerning how this vision of human-centric AI systems might look like.[132] Instead, it is up to the market players working in different fields to find various and specific solutions for particular industries and data spaces.

---

125. *Id.* at 42.

126. Artificial Intelligence Act Proposal, *supra* note 122, amend. 15, recital 4(a); *see also* Paulius Jurcys, *Human-Centric AI: The Missing Piece of the Debate on AI Networks*, MEDIUM (Oct. 10, 2023), https://medium.com/prifina/human-centric-ai-the-missing-piece-of-the-debate-on-ai-networks-264ff4eec408 [https://perma.cc/8QLG-9T6Q].

127. *See* EU AI Act, *supra* note 121, at 44.

128. Artificial Intelligence Act Proposal, *supra* note 122, amend. 15, recital 4(a).

129. See EU AI Act, *supra* note 121, at 2.

130. Artificial Intelligence Act Proposal, *supra* note 122, amend. 27, recital 9(a), amend. 213, art. 4(a).

131. *See* EU AI Act, *supra* note 121, at 44 ("[T]he purpose of this Regulation is to improve the functioning of the internal market and-promote the uptake of human-centric and trustworthy artificial intelligence (AI), while ensuring a high level of protection of health, safety, fundamental rights enshrined in the Charter, including democracy, the rule of law and environmental protection, against the harmful effects of AI systems in the Union and supporting innovation.").

132. *See id.* at 4.

### *F. Interim Conclusions*

Rapidly evolving data and AI models call for more coherent policy approaches. The current regulatory developments in the European Union demonstrate the market's need to make data more accessible and portable across platforms.[133] A human-centric approach to data and AI has emerged as a possible angle to address emerging technology pertaining to data and AI models.[134]

Although Article 20 of the GDPR provides a solid foundation for data portability, it also has several shortcomings and limitations. First, the term "data portability" remains ambiguously defined and lacks a cohesive interpretation.[135] Second, conflicting scopes of the right to data portability across different regulations make the situation even more complex.[136] Third, more clarity when it comes to data that is generated by the use of products in connection with the use of certain services is required. Fourth, from a technological standpoint, it is uncertain whether "data portability" signifies an effective data transfer or merely a singular, *in-situ* data access (e.g., on a device). Similarly, differences remain as to whether the right to data portability is effectuated by a one-time interaction or whether it mandates continuous, real-time data access.[137] Fifth, significant incongruences occur because of the different material scopes of European regulations.[138] For example, the Data Act is narrowly focused, encompassing only IoT data and excluding digital and online services.[139] In contrast, the DMA focuses predominantly on the dominion of gatekeepers in the digital space.[140] Sixth, although standardization and data interoperability are among the foundational elements of the portability right, it remains largely unregulated.[141] Seventh, from an individual user perspective, the exercise of data portability rights is cumbersome, manual, and time-consuming.[142] Finally, from the regulatory point of view, the question of "deciding who decides" remains pivotal. It is important to acknowledge and consider these existing uncertainties when examining possible paths forward.

---

133. For a broader discussion on the increasing role of data intermediaries, *see, e.g.*, Gabriele Carovano & Michèle Finck, *Regulating Data Intermediaries: The Impact of the Data Governance Act on the EU's Data Economy*, 50 COMPUT. L. & SEC. REV. 1, 2 (2023); Anne Josephine Flanagan & Sheila Warren, *Advancing Digital Agency: The Power of Data Intermediaries*, WORLD ECON. F. 1, 17 (Feb. 2022), https://www3.weforum.org/docs/WEF_Advancing_towards_Digital_Agency_2022.pdf [https://perma.cc/FV3V-5ZLL].

134. *See* Paulius Jurcys, Marcelo Corrales Compagnucci & Mark Fenwick, *The Future of International Data Transfers: Managing Legal Risk with A 'User-held' Data Model*, 46 COMPUT. L. & SEC. REV. 1, 7 (2022).

135. *See, e.g.*, Turner & Tanczer, *supra* note 7, at 4; Jurre Reus & Nicole Bilderbeek, *Data portability in the EU: An Obscure Data Subject Right*, IAPP (Mar. 24, 2022), https://iapp.org/news/a/data-portability-in-the-eu-an-obscure-data-subject-right [https://perma.cc/DF5T-XBT2].

136. Under the GDPR, the right to data portability is limited only to personal information "*provided*" by an individual to an online service, while under the Data Act, data portability covers not only personal data, but also datasets including a mix of personal and nonpersonal data generated by connected IoT devices. General Data Protection Regulation, *supra* note 20, at 45; Data Act, *supra* note 54, at 2.

137. *See, e.g.*, Paulius Jurcys, Chris Donewald, Jure Globocnik & Markus Lampinen, *My Data, My Terms: A Proposal for Personal Data Use Licenses*, 33 HARV. J.L. & TECH. DIG. 1, 9–12 (2020).

138. *See* discussion *supra* Sections II. A–II.D.

More generally, the current legal usage of data portability captures something important about a person's relationship with their own data. An individual's personal data is fluid, dynamic, and moveable. It has value to them, thus they should be the one that controls when and where it is moved. If an individual wants to use their data in another context, then they should be able to do so by their own choice. This notion has intuitive appeal: it connects to the idea of being able to retain what is important to an individual as they make changes to life projects and it emphasizes the importance of their data in those life projects.[143] Yet, it is worth exploring whether current legal usages—even in jurisdictions with relatively strong protections, like the European Union—go far enough in acknowledging the central importance of data portability in our lives and the public demand for more control. Or does the extant regulatory framework reflect and preserve an enterprise-centric view of these questions? In the next section, this Article considers technological developments in the market that offer tantalizing new possibilities for a more human-centric approach than the current regulatory scheme seems to support.

---

139. *See* discussion *supra* Section II.B.

140. *See* discussion *supra* Section II.C.

141. *See* Mark A. Lemley, Eric. E. Johnson & M. Christopher Riley, Stanford Interdisciplinary Working Group on Interoperability: Report and Preliminary Recommendations 7, (Apr. 27, 2023) (unpublished manuscript) (on file with SSRN), https://papers.ssrn.com/sol3/papers.cfm?abstract_id=4412862 [https://perma.cc/HX22-8G3T]; Kuebler-Wachendorff et al., *The Right to Data Portability*, *supra* note 18, at 265–66; Organisation for Economic Co-operation & Development [OECD], *Data Portability, Interoperability and Digital Platform Competition*, OECD (June 2021), http://oe.cd/dpic [https://perma.cc/SX99-8B9W].

142. *See, e.g.*, Turner & Tanczer, *supra* note 7, at 1 (arguing that data portability "is rendered meaningless without data subject's ability to exercise it in practice").

143. *See, e.g.*, Stefan Buehler, Ralf Dewenter & Justus Haucap, *Mobile Number Portability in Europe* 1 (Helmut-Schmidt-Universität, Universität der Bundeswehr Hamburg Working Paper, Diskussionspapier No. 41, 2005), http://hdl.handle.net/10419/23628 [https://perma.cc/X6ZP-9W3S].

## III. Technical Frameworks & Market Trends in Data Portability

The world is at a turning point where technology-driven developments in the market raise the possibility of a more human-centric approach in which individuals can have significantly greater dominion and control over their own data.[144] Technologies that make data collection and analysis easier and more effective, such as federated cloud architecture, edge computing, and machine learning technologies have matured to empower individuals with their data in new and previously unimaginable ways.[145]

The concepts of data access and data portability have evolved into three possible technological models, which will be introduced in the next three subsections. This Article will elaborate on the key features of these three approaches toward personal data governance and explore how each of the models affect and facilitate data portability and related concepts such as data access and data ownership, as well as the ramifications for data interoperability more generally. The first model reflects current EU regulatory thinking as described above, as well as similar regulatory schemes found elsewhere. The second and third models emerge "bottom-up" due to advancements in various technologies that can now be scaled commercially.

The comparison of three technological approaches from the perspective of how they treat user-generated data offers a glimpse of a paradigm shift. Namely, such a comparison of different data architectures clarifies how the technology and market moves away from traditional enterprise-centric models towards more human-centric data governance frameworks that recognize and respond to the public demand for greater individual control and dominion over personal data. It is important that regulators are cognizant of these trends and, if necessary, encourage them to align new regulatory frameworks with the ongoing technological and market developments. In short, there is a need to better align regulatory frameworks with the value of an individual's data in the digital age. That is, regulators must acknowledge the centrality of data in a post-digital transformation world and design regulatory frameworks that better reflect the meaning and significance of personal or user-generated data. This entails developing what this Article calls a "human-centric approach" that shifts from privacy as a super-qualified right built around formal but ultimately unsatisfactory notions of consent to "private-by-default" data framework in which the default position is for individuals to retain dominion over "their" data.[146]

---

144. *See* Fenwick & Jurcys, *supra* note 9; Paulius Jurcys, Christopher Donewald, Mark Fenwick, Markus Lampinen, Vytautas Nekrošius & Andrius Smaliukas, *Ownership of User-Held Data: Why Property Law is the Right Approach*, HARV. J.L. & TECH. DIGEST 1, 7 (Sept. 21, 2021), https://jolt.law.harvard.edu/digest/ownership-of-user-held-data-why-property-law-is-the-right-approach [https://perma.cc/F6W3-EUAW].

145. Fenwick & Jurcys, *supra* note 9, at 391.

### *A. A Consent & Opt-Out Model*

The first approach to personal data portability is based on the idea that individuals should have control over their data and be able to decide how, when, and where it is used and with whom it is shared. The consent and opt-out model is essentially enshrined in the GDPR, as well as similar laws elsewhere (like the Californian CCPA).[147] The technical architecture that reflects this regulatory model is illustrated in Figure 1, which places emphasis on the idea that users have the power to control their data by giving consent and opting out of certain data collection, processing, and use practices adopted by the business with which an individual interacts. In other words, in this user-controlled data governance model, the notion of control over one's data essentially means the right to opt out.[148]

After the GDPR was adopted, hundreds of companies providing data privacy compliance services for enterprises emerged.[149] Most of those data compliance service providers are based in Europe and the United States.[150] These companies can be categorized into two groups. The first group primarily serves other enterprises in managing consumer data and establishing data privacy compliance programs. The second group of service providers helps individual consumers to manage their data rights, permissions, and consents. This second group of companies is sometimes referred to as "data fiduciaries," which should, in principle, mean that they act in the best interest of consumers when helping them exercise their data rights.[151]

---

146. Paulius Jurcys, Mark Fenwick & Souichiro Kozuka, "Private-By-Default": A Data Framework for the Age of Personal AI 9 (Nov. 30, 2024) (unpublished article) (on file with SSRN), https://papers.ssrn.com/sol3/papers.cfm?abstract_id=4839183 [https://perma.cc/MP4G-ZL4V].

147. Fenwick & Jurcys, *supra* note 9, at 391.

148. California Consumer Privacy Act of 2018, CAL. CIV. CODE §§ 1798.100–1798.199.100. In certain jurisdictions such as California, users can also request for companies not to sell their data to third parties or to delete all the data collected about that user. *See id.* § 1798.105.

149. For an academic discussion, *see* Jack M. Balkin, *The Fiduciary Model of Privacy*, 133 HARV. L. REV. F. 11, 12–13 (2020) (with further references); Heiko Richter & Peter R. Slowinski, *The Data Sharing Economy: On the Emergence of New Intermediaries*, 50 INT'L REV. INTELL. PROP. & COMPETITION L. 4, 8 (2019); Joss Langford, Antti 'Jogi' Poikola, Wil Janssen, Viivi Lähteenoja & Marlies Rikken, *Understanding MyData Operators*, MYDATA 1, 54–62 (Mar. 16, 2022), https://mydata.org/publication/understanding-mydata-operators/ [https://perma.cc/DX8V-U4SE].

150. This is because of the "Brussels effect" of the GDPR and Californian CCPA which provided incentives for such new services to emerge and help consumers manage their data rights. *See* Anu Bradford, *The Brussels Effect*, 107 Nw. U.L. Rev. 1, 3, 24 (2015).

As for data portability, a user's ability to consent and control means that users have two options: (i) to request Service Provider A to "give back" the data it has collected about the user and bring such data to Service Provider B, or (ii) to instruct that Service Provider A transfers the data about the user to another service provider designated by the user.[152] Consumer-facing consent management companies offer their services to help individuals exercise this right.[153]

Figure 1. Data Portability in Consent & Opt-Out Model

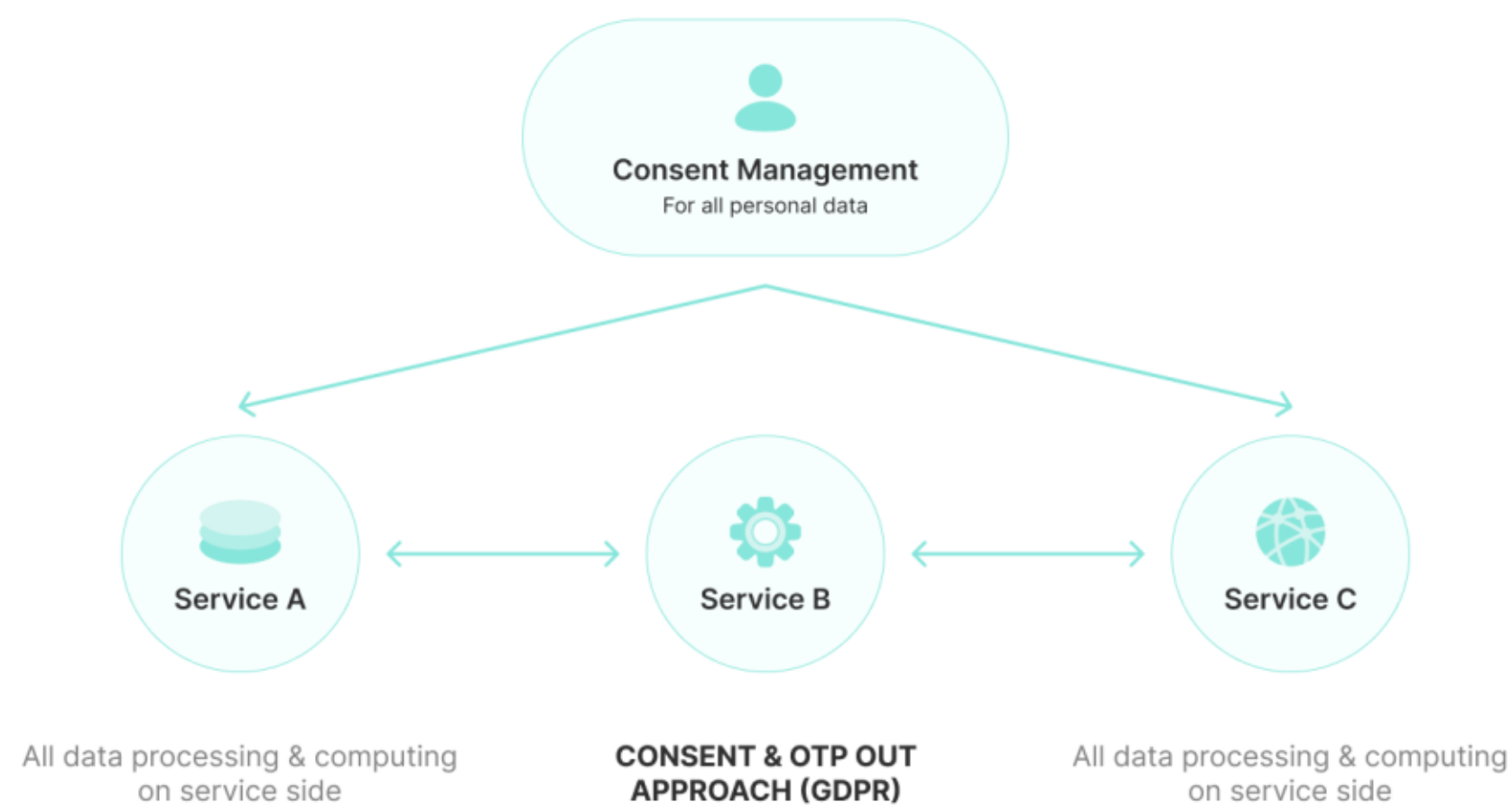

151. Noelle Wilson & Amanda Reid, *Data Controllers as Data Fiduciaries: Theory, Definitions & Burdens of Proof*, 95 U. COLO. L. REV. 175, 181 (2024). For example, an Israeli-based mine helps individuals to 'own' the data by giving them easy tools to control who can access their data and revoke permissions. *Transparency*, SAYMINE, https://www.saymine.com/transparency [https://perma.cc/NE6D-QL5S] (last visited Feb. 3, 2025); *Contact*, SAYMINE, https://www.saymine.com/contact [https://perma.cc/NVW5-G74J] (last visited Feb. 3, 2025).

152. *See, e.g.*, General Data Protection Regulation, *supra* note 20, at 45.

153. Dan Frechtling, *The Mismanagement of User Consent Data and Its Consequences*, IAPP, (Mar. 9, 2023), https://iapp.org/news/a/the-mismanagement-of-user-consent-data-and-its-consequences [https://perma.cc/D2WE-WU3M]. Although it has been quite difficult in practice because service providers impose high identity verification requirements on individual consumers which complicates the exercise of the right of data portability via third-party "authorized agents." *See* PAULIUS JURCYS & MARK LAMPINEN, PRINCIPLES OF DATA PRIVACY IN CALIFORNIA: STUDY OF INDUSTRY REACTIONS AND COMMENTS TO THE PROPOSED CCPA REGULATIONS AND USER-CENTRIC PERSPECTIVES 40 (2020).

More generally, the consent and opt-out data governance model aims to ensure that the user's data is portable among and between the services, which benefits the user.[154] To implement this data portability framework, each consumer-facing service provider must maintain its own data platform to manage all its customers' data requests.[155] However, in this model, individuals have only limited agency over their personal data. Consumer data access and portability rights are limited to the right to instruct and request a copy of their data.[156] In this consent and opt-out model, consumers do not get a personal data storage solution where their personal data is aggregated in a place that could be under their direct control—they need to figure that out by themselves.[157]

The consent and opt-out data governance model is essentially enshrined in the GDPR and Californian CCPA, and the technical architecture for this model is mostly oriented around several aspects of the consumers' right to opt out of various data collection practices by service providers.[158] One of the main advantages of this model is that it empowers individuals to control how services access, collect, use, and share consumer data.[159] It provides a *sense* of control and *sense* of ownership of data by allowing users to instruct service providers to behave in a certain way regarding personal data.[160] Some consumer data rights management platforms do quite a good job of promoting consumer data literacy. For instance, consumers might be provided with a dashboard that shows hundreds of companies that have some digital footprints of consumers based on past interactions.[161] Consumers could then rely on easy-to-use tools to exercise their GDPR and CCPA rights and access their "privacy score," which reflects to what

---

154. *See infra* p. 125.

155. *See* PETER SWIRE & DEBRAE KENNEDY-MAYO, U.S. PRIVATE-SECTOR PRIVACY, 69 (2d ed. 2018).

156. *See, e.g.*, General Data Protection Regulation, *supra* note 20, at 45.

157. *See, e.g.*, *id.*

158. *See id.*; California Consumer Privacy Act of 2018, CAL. CIV. CODE § 1798.120(a)(1). The CCPA stands out by entrenching consumers' right to opt out from sales of data and establishing a mandatory data broker registry. *See Data Broker Registry*, STATE CAL. DEP'T JUST., https://oag.ca.gov/data-brokers [https://perma.cc/J9JF-SUAJ] (last visited Feb. 3, 2025).

159. *See* Janis Wong & Tristan Henderson, *The Right To Data Portability In Practice: Exploring the Implications of the Technologically Neutral GDPR*, 9 INT'L DATA PRIV. L. 173, 174 (2019).

160. *See* Rubinfeld, *supra* note 49, at 173.

161. *See, e.g.*, Paulius Jurcys, *What Is the Value of Your Data?*, MEDIUM 1, 5 (Sept. 5, 2019), https://medium.com/prifina/what-is-the-value-of-your-data-9341cd019b4d [https://perma.cc/Q7SJ-WDWC] (providing an illustration of data rights management platform SayMine's consumer-facing dashboard).

degree a consumer has exercised their data rights vis-à-vis service providers.[162]

This consent and opt-out model has several important implications for data interoperability. First, companies (especially technology giants) are less motivated to create truly interoperable environments because there is little economic incentive for them to do that.[163] Therefore, data access requests are performed by sending consumers files in JSON format, which, although human-readable, is only useful for software developers and data scientists because they have the necessary data literacy skills to understand data in JSON.[164] In practice, this consent and opt-out model means that an individual's data ends up being locked in the hands of service providers.[165] Individual users do not *have* any data; they can only manage how they gain access to the data service providers have already collected about them.[166] In turn, they can exercise only limited opt-out rights as to how such service providers use and share some of that data with third parties.

Second, this model has significant scalability challenges for data security and interoperability. With three services, there are three possible connections; with four services, there are six possible connections, and so on.[167] Therefore, the more services using an individual's personal data, the higher the risk of data security breaches becomes (see Figure 2 below). Communication among systems is subject to a combinatorial explosion.[168] In other words, as more systems are added to the overall setup, the lines of connections are subject to a rapidly accelerating risk of data breaches. Scalability overheads thus increase dramatically, which very quickly comes at the expense of feasibility. For example, with only seven connected systems, twenty-one connections are required.[169] If a US or EU consumer has eighty service

162. *See, e.g.*, *Your Guide to Privacy Scores*, PRIVACYMONITOR, https://www.privacymonitor.com/score/ [https://perma.cc/H62F-MG67] (last visited Feb. 14, 2025).

163 *See* Miriam Reisman, *EHRs: The Challenge of Making Electronic Data Usable and Interoperable*, 42 P&T 572, 573 (2017).

164. *See* Lindsay Rowntree, *Remember Why the GDPR Exists: Consumers Need Friendly GDPR Functions*, EXCHANGEWIRE (Sept. 28, 2018), https://www.exchangewire.com/blog/2018/09/28/remember-why-the-gdpr-exists-consumers-need-friendly-gdpr-functions/ [https://perma.cc/KBY4-ET7V].

165. *See id.*

166. *See id.*

167. *See infra* Figure 2; *see also* General Data Protection Regulation, *supra* note 20, recital 6 (discussing the increasing scale of collection and sharing of personal data).

168. *See infra* Figure 2.

169. *See id.*

providers on average—which could be tantamount to the number of apps on their phones—how could data portability be effectively realized?[170]

Figure 2. Scalability Problems in the Consent & Control Model

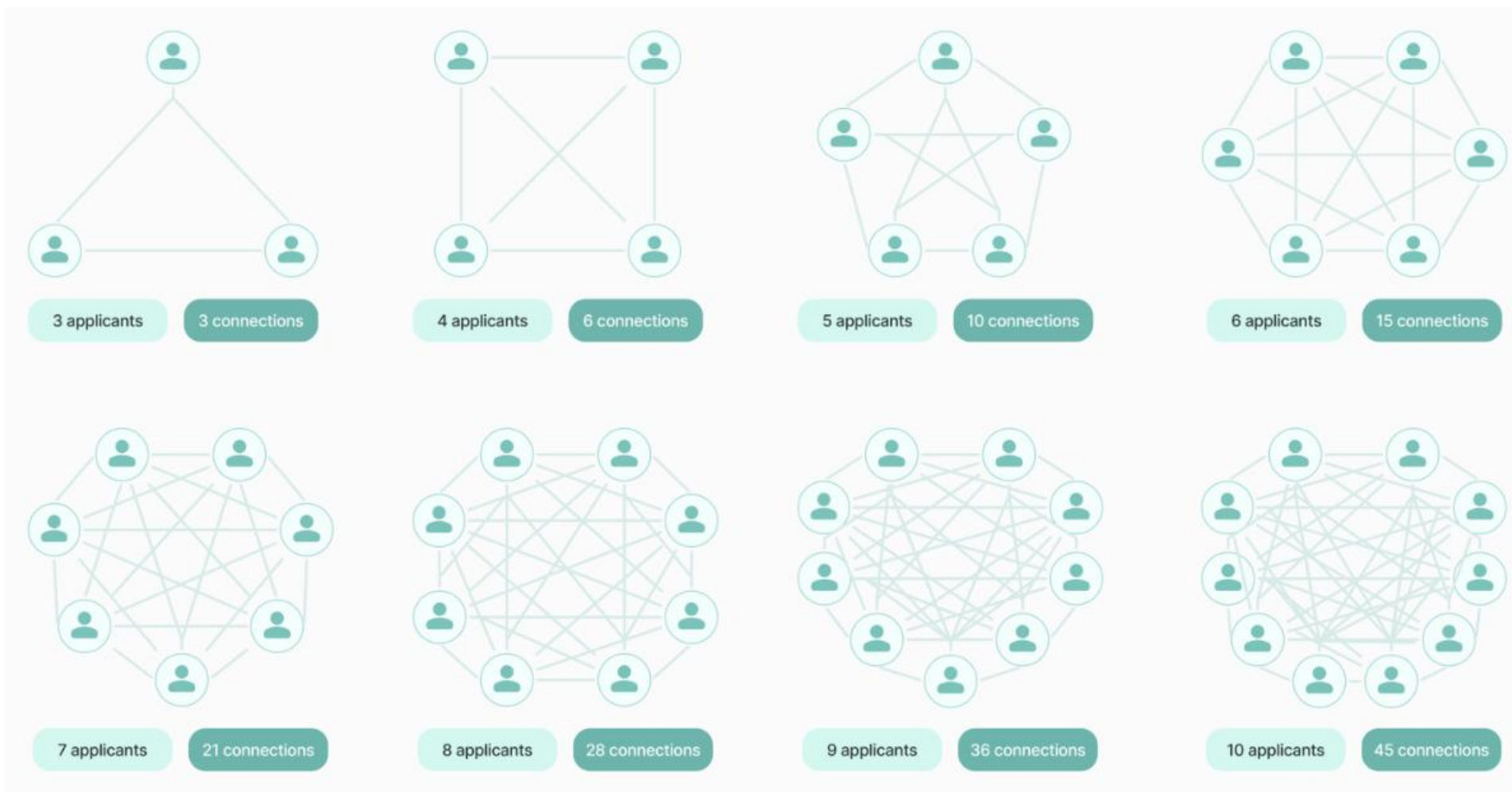

Third, all data processing and computing is done on the service provider's side.[171] Thus, the service provider is responsible for processing and analyzing the data and ensuring that the value from user-generated data is captured on the service provider's side rather than the individual's.[172] This can lead to potential security and privacy concerns, as the service provider may not be able to guarantee a level of protection and control that would satisfy individual users trying to port their data between different service providers.

170. *See Mobile App Download Statistics & Usage Statistics*, BUILDFIRE (Dec. 31, 2024), https://buildfire.com/app-statistics/ [https://perma.cc/K2SK-MAMT].

171. *See, e.g.*, General Data Protection Regulation, *supra* note 20, at 32–33. Because the data is collected and processed in centralized servers of service providers, the currently existing global data system is often described as "siloed" and highly guarded by each service provider. *See* Scott Robinson, *What Are Data Silos and What Problems Do They Cause?*, TECHTARGET (July 2024), https://www.techtarget.com/searchdatamanagement/definition/data-silo [https://perma.cc/4K9E-33PV].

172. *See* CARISSA VÉLIZ, PRIVACY IS POWER: WHY AND HOW YOU SHOULD TAKE BACK CONTROL OF YOUR DATA 7 (2021).

### *B. Enterprise-Centric Approaches*

A second technical approach to data portability and personal data governance revolves around enterprise-centric solutions. This section discusses two projects aimed to unlock data portability primarily *between* enterprises. One of them—Gaia-X—is facilitated by the governments of several European Union member States;[173] the other is spearheaded by a private entity firm, Inrupt.[174]

#### 1. GAIA-X

GAIA-X is an European Union-wide initiative to build an infrastructure and data ecosystem according to European values and standards.[175] GAIA-X was envisioned as data infrastructure and an open digital ecosystem "initiated by Europe, for Europe."[176] It aims to support European companies' global competitiveness and promote "European data sovereignty and data availability."[177] In particular, the GAIA-X project is aligned with the European Data Strategy and aims to provide businesses with an easy, secure, and safe way to access an infinite amount of high-quality industrial data.[178]

The GAIA-X project was initiated by the governments of Germany and France and aims to address many concerns with data collection, data processing, and data transfers.[179] GAIA-X is very much a work in progress: while there is a consensus on the technical architecture of GAIA-X, it is far from becoming a norm.[180] By and large, the architecture of GAIA-X goes far beyond the highly centralized cloud approach.[181] It supports the following principles: security by design, privacy by design, federation, distribution, and decentralization of data.[182] Moreover, there are guidelines for user friendliness like improving the user experience by reducing friction.[183] The relationships among stakeholders—consumers, data owners, and service and node providers—are clearly defined.[184] Figure 3 below offers a high-level overview of the technical architecture of GAIA-X.

---

173. *Gaia-X Explained*, GAIA-X HUB GER., https://gaia-x-hub.de/en/gaia-x-explained/ [https://perma.cc/ND7W-UV59] (last visited Feb. 14, 2025).

174. *See* INRUPT, https://www.inrupt.com/ [https://perma.cc/8MBV-BFE4] (last visited Jan. 31, 2025).

175. *See Gaia-X Explained*, *supra* note 173.

176. *What Is GAIA-X and What Do I Need to Know?*, SQUIRE PATTON BOGGS 1, 2 (Nov. 2022), https://www.squirepattonboggs.com/-/media/files/insights/publications/2020/11/what-is-gaia-x-and-what-do-i-need-to-know/what-is-gaia-x-and-what-do-i-need-to-know.pdf. [https://perma.cc/4AFW-ZMD6].

177. *Optimizing the European Construction Industry: Making Construction More Digital*, HOCHSCHULE HOF CAMPULS, https://campuls.hof-university.com/science-research/optimierung-der-europaeischen-bauwirtschaft-bauen-soll-einfacher-und-digitaler-werden-2/ [https://perma.cc/Y6L4-AQTJ] (last visited Feb. 14, 2025).

178. *See* Robert Goia, *Gaia-X: The Democratic and Legal Aspects of the Visionary European Cloud Project*, (July 13, 2022), https://gaia-x.eu/news-press/gaia-x-the-democratic-and-legal-aspects-of-the-visionary-european-cloud-project/ [https://perma.cc/Z6LQ-CQN9].

179. *See* Hubert Tardieu, *Role of Gaia-X in the European Data Space Ecosystem*, *in* DESIGNING DATA SPACES 41 (Boris Otto, Michael Ten Hompel & Stefan Wrobel eds., 2022).

180. *See id.*

Figure 3. GAIA-X Architecture[185]

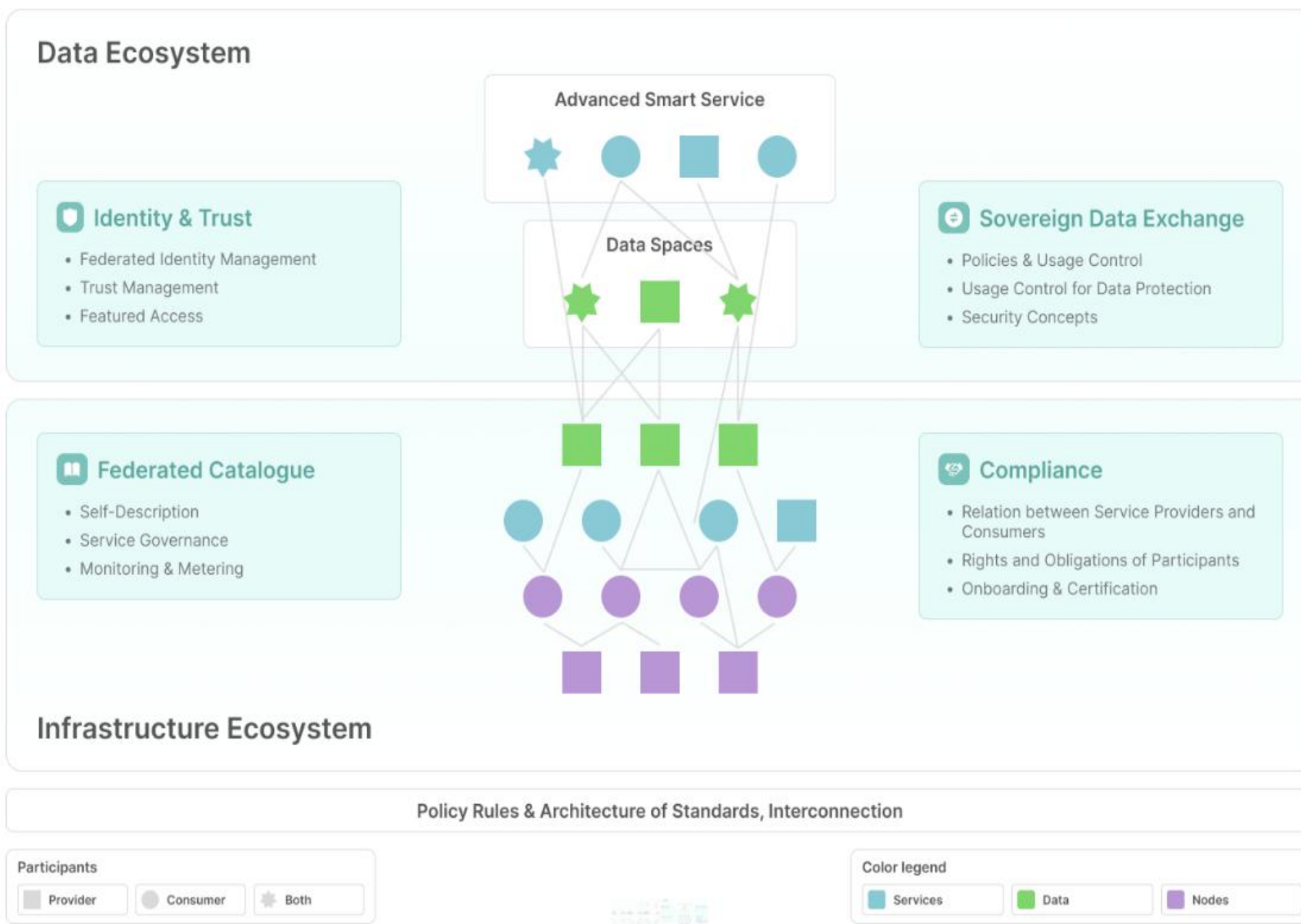

The GAIA-X project rests upon the following four high-level principles that represent the core values of the GAIA-X architecture[186]:

1. *Openness and transparency*: GAIA-X technologies (the specifications and documentation) are all publicly available.[187]

181. DE-CIX MANAGEMENT, GUNTER EGGERS, BERND FONDERMANN, GOOGLE GERMANY GMBH, BERTHOLD MAIER, KLAUS OTTRADOVETZ, JULIUS PFROMMER, RONNY REINHARDT, HANNES ROLLIN, ARNE SCHMIEG, SEBASTIAN STEINBUß, PHILIPP TRINIUS, ANDREAS WEISS, DR. CHRISTIAN WEISS & DR. SABINE WILFLING, GAIA-X: TECHNICAL ARCHITECTURE 4 (2020).

182. *Id.*

183. *Id.*

184. *See id.* at 7.

185. *Id.* at 5.

186. *Id.* at 3–4.

187. *Id.* at 3.

The aim is to involve as many public and private stakeholders as possible; therefore, open-source licenses are used to facilitate the distribution of the technologies and the emergence of uniform standards.[188]

2. *Interoperability*: GAIA-X architecture provides an agnostic technical foundation so that all participants can interact with each other without relying on specific technology implementations.[189]
3. *Federated systems*: GAIA-X aims to support decentralization and distribution by specifying federated systems of autonomous actors who follow a common set of standards, frameworks, and legal rules.[190]
4. *Authenticity and trust*: GAIA-X architecture goes beyond the authority of a single organization. As a network of multiple actors, GAIA-X relies on identity management systems with mutual authentication and selective disclosures.[191] Such mutual dependency of actors should nurture mutual trust among them and encourage the growth of a secure digital ecosystem.[192]

GAIA-X aims to foster the development of ecosystems for infrastructure and data services.[193] The GAIA-X ecosystem is enabled by interoperability on a technical and organizational level, allowing seamless integration and use of offerings across vendors.[194] GAIA-X specifically addresses the following topics to facilitate interoperability in ecosystems: identity and trust management, discovery, standards for interoperability (i.e., the architecture of standards), enforceable usage policies, contracting between the data provider and data consumer, and monitoring and metering.[195]

There are two main layers of the GAIA-X ecosystem: the infrastructure layer and the data layer.[196] The infrastructure ecosystem consists of services and necessary infrastructure components to store, transfer, and process data.[197] The "federated GAIA-X concept provides services across multiple [p]roviders and [n]odes of the ecosystem."[198] Infrastructure services can range from low-level services like bare metal computing to sophisticated offerings, like high-performance computing.[199] Robust connectivity services ensure secure performance data exchange between different data providers and services.[200]

---

188. *See id.*

189. *See id.* at 4.

190. *Id.*

191. *See id.*

192. *Id.* There are approximately three hundred member organizations within the GAIA-X community. *Who Is Involved in Gaia-X*, *in FAQ*, GAIA-X, https://gaia-x.eu/faq/ [https://perma.cc/2JGY-J8MJ] (last visited Jan. 31, 2025). *Members Directory*, GAIA-X, https://gaia-x.eu/community/members-directory/ [https://perma.cc/QJ6S-2NG7] (last visited Jan. 31, 2025).

193. DE-CIX MANAGEMENT ET AL., *supra* note 181, at 25.

194. *Id.*

195. *Id.*

196. *Id.*

Currently, there are around three hundred governmental and private business entities—from tech giants to start-ups—exploring various use-cases that could be built using GAIA-X architecture.[201] Such use-cases range from various models for specific sectors like agriculture, education, energy, manufacturing, medical and wellness data, supply chain and more, as well as cross-sectoral applications like smart cities.[202] One unifying feature for all stakeholders operating in the GAIA-X framework is the high compliance thresholds to which all participants are obliged to adhere.[203] Such mandatory compliance requirements concern information security and data protection that are expected to enhance the mutual trust among all participants and in the GAIA-X framework itself.[204]

With regard to data protection, the GAIA-X framework relies upon the standards enshrined in the GDPR.[205] To facilitate the development of a trusted GAIA-X environment and effectively utilize existing standards, data processing parties can declare themselves subject to two mechanisms to voluntarily underpin their compliance with GDPR requirements: (1) codes of conduct pursuant to Articles 40 and 41 of the GDPR and (2) certifications pursuant to Articles 42 and 43 of the GDPR.[206] The data processors can use these standards while also taking advantage of legal incentives under the GDPR.[207] The framers of GAIA-X do not believe that there will be one overarching standard that verifies compliance with all possible and applicable GDPR requirements.[208] On the contrary, it is expected that the GDPR standards relevant to GAIA-X—both Codes of Conduct and

197. *Id.*
198. *Id.*
199. *Id.*
200. *Id.*
201. *See* GAIA-X, *supra* note 192.
202. *See id.*; *Community*, GAIA-X, https://gaia-x.eu/community/ [https://perma.cc/3AWQ-JNWU] (last visited Jan. 31, 2025).
203. *See* DE-CIX MANAGEMENT ET AL., *supra* note 181, at 31.
204. *See id.* at 30.
205. *See id.* at 32.
206. *Id.*
207. *Id.*
208. *Id.* at 33.

Certifications—will either address specific market sectors or specific processing activities.[209]

### 2. A Personal Data Wallet (Pod) Model

The second technological approach to data portability between enterprises is the "personal data wallet" model.[210] This model is an evolution of the consent and control approach to personal data discussed in Section III. A. The difference, however, is that in this model, each user is provided with a personal data wallet (sometimes called "personal data pod" or "personal data store") with a unique user ID.[211] The user's data wallet is the center of interaction between connected applications or services and the individual user.[212] The user is, therefore, in control of their personal data and can exercise such control by allowing service providers to access their data.[213] The personal data pod or wallet is provided by an intermediary service provider like Inrupt to enterprises such as CNN, BBC, or a regional government that conducted pilots to store and manage their customer data in the pod.[214]

One of the core features of this personal data pod approach is the separation of data from apps.[215] A personal data wallet provides a place for organizations to merge all the data they have about each individual customer in separate containers.[216] In other words, a personal data wallet is similar to a warehouse where organizations can store and access their customers' data.[217] This data framework relies on a universal data infrastructure and standard interface, which helps companies access customer data with customers' consent (in line with their privacy agreements) and minimizes the use of development resources to build new applications and services.[218] When data is organized in every customer's wallet, customers gain more visibility into how organizations use and access their personal information; individuals can also manage how their data is shared with organizations.[219] For organizations, this customer-centric data architecture should simplify compliance with data privacy regulations and offer a lower-cost method to personalize services.[220]

---

209. *Id.*

210. *See* Ron Miller, *Tim Berners Lee's Startup Inrupt Releases Solid Privacy Platform for Enterprises*, TECHCRUNCH (Nov. 8, 2020, 9:01 PM), https://techcrunch.com/2020/11/08/tim-berners-lees-startup-inrupt-releases-solid-privacy-platform-for-enterprises/ [https://perma.cc/SH7C-YPNN].

211. *See* Osmar Olivo, *Web 3.0 Doesn't Need a Blockchain Revolution*, INFOWORLD (Jan. 19, 2023), https://www.infoworld.com/article/3685572/web-3-doesnt-need-a-blockchain-revolution.html [https://perma.cc/N5VK-9M7F].

212. *See id.*

213. *See id.*

214. *See* Miller, *supra* note 210. The most prominent company offering personal data pod infrastructure is Inrupt and its enterprise-focused solution Solid. *See id.* (referencing INRUPT, *supra* note 174).

215. *See* Washington Post Live, *Sir Tim Berners-Lee Says His Company Solid Will Give Users Control of Their Data*, YOUTUBE (Mar. 6, 2019), https://www.youtube.com/watch?v=eJ6IrWc7Wt4; *see also infra* Figure 4.

216. *What Do Personal Data Stores Mean for Privacy Regulations, Digital Identity, and Customer Trust*, INRUPT (Feb. 28, 2023), https://www.inrupt.com/blog/what-do-personal-data-stores-mean-for-privacy-regulations-digital-identity-and-customer-trust [https://perma.cc/NB8W-XC9E].

Figure 4. The Personal Data Wallet/Pod Model

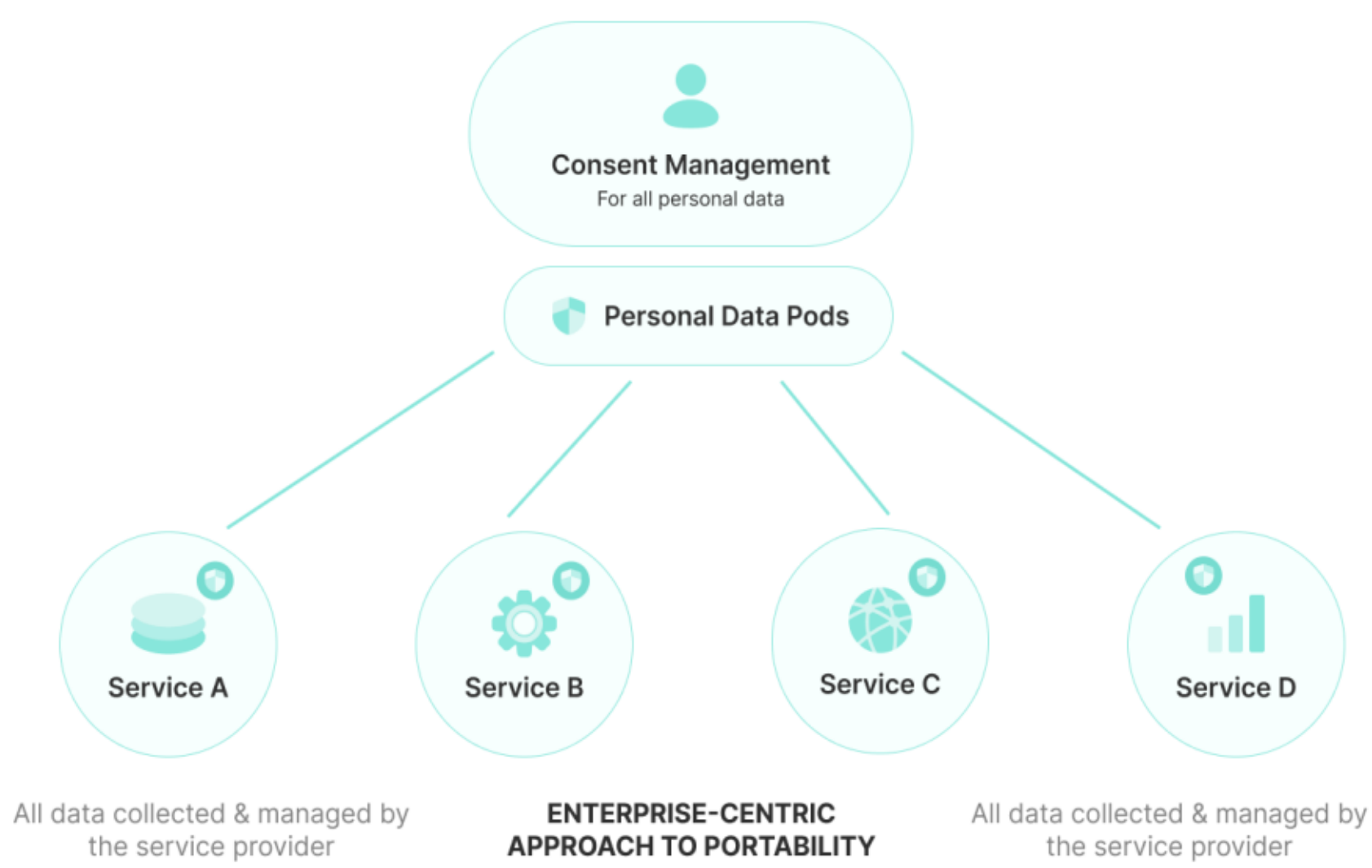

There are two publicized case studies about adopting the personal data pod model implemented by large organizations. The first one was a test

217. *See* Emmet Townsend, *Engineering VP Unveils Data Warehouse Management Benefits*, SOLS. REV. (Mar. 2, 2022), https://solutionsreview.com/data-management/engineering-vp-unveils-data-warehouse-management-benefits/ [https://perma.cc/C65M-VXEB].

218. *See id.*

219. *What Do Personal Data Stores Mean for Privacy Regulations, Digital Identity, and Customer Trust*, *supra* note 216; *Digital Flanders Reconnects Citizens with Their Data Through Inrupt's Solid Server*, INRUPT (Sept. 15, 2022) [hereinafter *Digital Flanders*], https://www.inrupt.com/blog/digital-flanders-reconnects-citizens-with-their-data-through-inrupts-solid-server [https://perma.cc/K9KQ-M6QL].

220. Townsend, *supra* note 217; *What Do Personal Data Stores Mean for Privacy Regulations, Digital Identity, and Customer Trust*, *supra* note 216.

case developed by the BBC, which provided personal data pods to selected users of BBC services.[221] Participating users were able to collect historical data on the content viewed or listened to on the BBC and other entertainment platforms like Spotify.[222] These historical profiles were used to help BBC customize the content for each pod owner.[223] The second project was implemented by Inrupt and the Flemish government to build a service called "My Citizen Profile," where every citizen of Flanders was given a data pod from birth.[224] Through their pods, the Flanders citizens should be able to update every government department in one place.[225] For example, Flanders citizens can input a change of address or the birth of a child.[226] They can even share their work history with potential employers or register a new company.[227]

Undoubtedly, this enterprise-centric data pod approach is an important step forward in the realm of personal data governance, especially compared to the consent and opt-out model.[228] A personal data pod not only offers more agency and control for individuals over their personal data and personal profiles, but it also allows the physical pod ecosystem to start collecting copies of their own data from various service providers.[229] As for data privacy and "control," the personal data pod model provides individuals possibilities to express *prior* consent to allow service providers access to user data in the data wallet rather than an ex-post ability to opt out of tracking, which is the essence of the consent and opt-out model.[230]

---

221. *The BBC Uses Inrupt's Solid Server to Deliver Viewers a Personalized but Private "Watch Party" Experience*, INRUPT (Oct. 27, 2022) [hereinafter INRUPT, *BBC*], https://www.inrupt.com/blog/bbc-delivers-personalized-private-viewing-parties-with-inrupt [https://perma.cc/6L9F-8EHL].

222. *Id.*

223. *How Media and Publishing Companies Can Embrace Data Transparency for Future Success*, INRUPT (Jan. 30, 2023) [hereinafter *How Media and Publishing Companies*], https://www.inrupt.com/blog/how-media-and-publishing-companies-can-embrace-data-transparency-for-future-success [https://perma.cc/ZG4U-RZZT].

224. *Digital Flanders*, *supra* note 219; James Shackell, *Rage Against the Machine: How the Inventor of the Web Is Trying to Save It*, ROLLING STONE (Sept. 16, 2022), https://au.rollingstone.com/culture/culture-features/web-rage-against-machine-42845/ [https://perma.cc/6PS5-FFVF].

225. Shackell, *supra* note 224.

226. *Id.*

227. *Id.*

228. *See Digital Flanders*, *supra* note 219 (describing Solid as "the only Enterprise-grade service in the world"); Townsend, *supra* note 217; *How Media and Publishing Companies*, *supra* note 223.

229. *See* Shackell, *supra* note 224.

230. *See id.*

Conceptually, the notion of a personal data pod is quite appealing. One of the advantages of this approach is that it could lead to a more comprehensive solution for personal data management and protection. The personal data pod may serve as a central hub for all the user's personal data, and allow the user to share their data with different services in a controlled and secure manner.[231] From a normative perspective, however, it is unclear whether this model solves the data privacy problem or augments it. The creators of personal data pod infrastructure acknowledge that users will have multiple pods created by various organizations and service providers.[232] This begs the question of whether risks emanating from data collections in siloed environments are mitigated as personal data and users' digital profiles will be further replicated in every additional personal data pod created for every consumer. While it is true that the personal data pod infrastructure brings services closer to the individual customer, creating additional pods with different service providers multiplies the number of interactions with every user's personal data.

More importantly, the entire personal data pod infrastructure is based on the premise that the user's personal data that is stored in the data pod is *not private*. Rather, the starting point for the entire data wallet infrastructure appears to be based on the notion that data stored in pods is *accessible by default*.[233] From a consumer privacy interest perspective, the personal data pod framework is primarily tailored for *enterprises* to handle their customers' personal data.[234] In other words, while it is true that users "own" their data in their data pod, the primary goal is to create a customer data interoperability framework for service providers, not consumers.[235] Uniform interoperability standards add value for content producers and distributors on the internet, such as news portals or content streaming service providers.[236] Hence, *who* determines the standards of data interoperability may be in question. In the personal data pod framework, data interoperability

---

231. *See id.*

232. *Ruben Verborgh on Data & Privacy*, IMEC, https://www.imec-int.com/en/imec-magazine/imec-magazine-january-2019/back-to-the-future-how-we-will-regain-control-of-our-personal-data [https://perma.cc/EWB7-F2PH] (last visited Jan. 31, 2025).

233. *See* Olivo, *supra* note 211, (discussing the access-by-default approach to data).

"With web-native solutions such as Solid, data is distributed. This means that regardless of where data is physically stored, it is connected to the person it describes, and the data is interoperable across systems. People are able to revoke access to most classes of data if they choose, but there is also support for cases where access must be granted to certain entities for compliance and governance reasons." *Id.*

234. *See, e.g.*, INRUPT, *BBC*, *supra* note 221; *see also* Townsend, *supra* note 217.

235. *See, e.g.*, INRUPT, *BBC*, *supra* note 221.

236. *See, e.g.*, *id.*

turns out to be quite centralized, and interoperability standards are established in a "top-down" fashion by the major technology companies or industry players running these personal data pod platforms.[237] From the regulatory perspective, this raises concerns about whether standards-setting functions should be bestowed upon certain public agencies or authorities.[238] Another alternative would be to develop some open-source standards determined by the community (a bottom-up approach).[239]

### *C. A Human-Centric Approach to Personal Data*

A third approach to personal data governance is a human-centric (or "user-held") data model. It has similarities to the first two models, but it pushes the discussion about data portability and interoperability forward in several new directions. The human-centric approach to data aims to empower individuals with their data by organizing user-generated data across platforms on the user's side and ensuring that the user's data is *private-by-default*.[240] In other words, a human centric approach to personal data refers to an infrastructure that is built around an individual consumer collecting data from various sources in their own personal data environment. Similar to enterprise-centric models described above, each individual's data layer is separate from applications in a human centric data framework.

The human-centric, user-held data model has two defining features.[241] First, each individual user has their personal data environment where they may collect data from different sources such as digital platforms, personal wearable and other IoT devices, data from mobile applications, and online services.[242] For example, a user may collect location data from Google Maps, shopping history from Amazon, or data generated while using online entertainment apps, among

237. *See* Townsend, *supra* note 217.

238. For a general overview, *see* Giuseppe Borgogno & Oscar Colangelo, *Open Banking and the Ambiguous Competitive Effects of Data Portability*, CPI ANTITRUST CHRONS., 1, 3–4 (Apr. 2021), https://papers.ssrn.com/sol3/papers.cfm?abstract_id=3826444 [https://perma.cc/X92K-8ZC8].

239. *See, e.g.*, Markus Lampinen & Paulius Jurcys, *Prifina Comments on the Regulation of the European Parliament and of the Council on Harmonised Rules on Fair Access to and Use of Data (Data Act)*, PRIFINA 1, 9 (June 1, 2022), https://papers.ssrn.com/sol3/papers.cfm?abstract_id=4110462 [https://perma.cc/2TJE-GBXX].

240. Fenwick & Jurcys, *supra* note 9, at 391.

241. *Id.*; *see e.g.*, *Our Story (So Far)*, PRIFINA, https://www.prifina.com/story.html [https://perma.cc/JAA5-UH26] (last visited Jan. 22, 2025) (building a human-centric data ecosystem where each user's data is private by default).

242. Fenwick & Jurcys, *supra* note 9, at 391.

others. Each individual's personal data environment contains an embedded software robot that unifies data from these different data sources.[243]

Second, the user-held data model is built on an open-source API infrastructure that allows anyone to develop intelligent applications capable of correlating different sets of personal data and providing new insights unique to each individual.[244] For instance, an app could enable users to see how their heart rate (measured by smart wearable devices like a Fitbit) correlates with the types of movies they watch on Netflix. From an ecosystem perspective, the separation of the data layer from the application layer ensures that individuals can easily create personal data apps for their own use. Just as anyone can create their own website or blog in the Web 2.0 environment, similar possibilities will arise for personal data applications in this new model.[245]

In a human-centric data framework, an individual's data stays on the individual user's side.[246] Based on their own preferences and needs, an individual can choose to install relevant apps into their own private data environments and "activate" their private data. In practice, this means that individuals do not lose control over their data. Instead, an individual's data is private-by-default, and AI-powered applications run locally, in a federated manner, and in the user's private data environment.[247]

243. *Id.*
244. *Id.*
245. *See id.*
246. *Id.*
247. *Id.* For a visual illustration, *see infra* Figure 5.

Figure 5. A Human-Centric Data Model

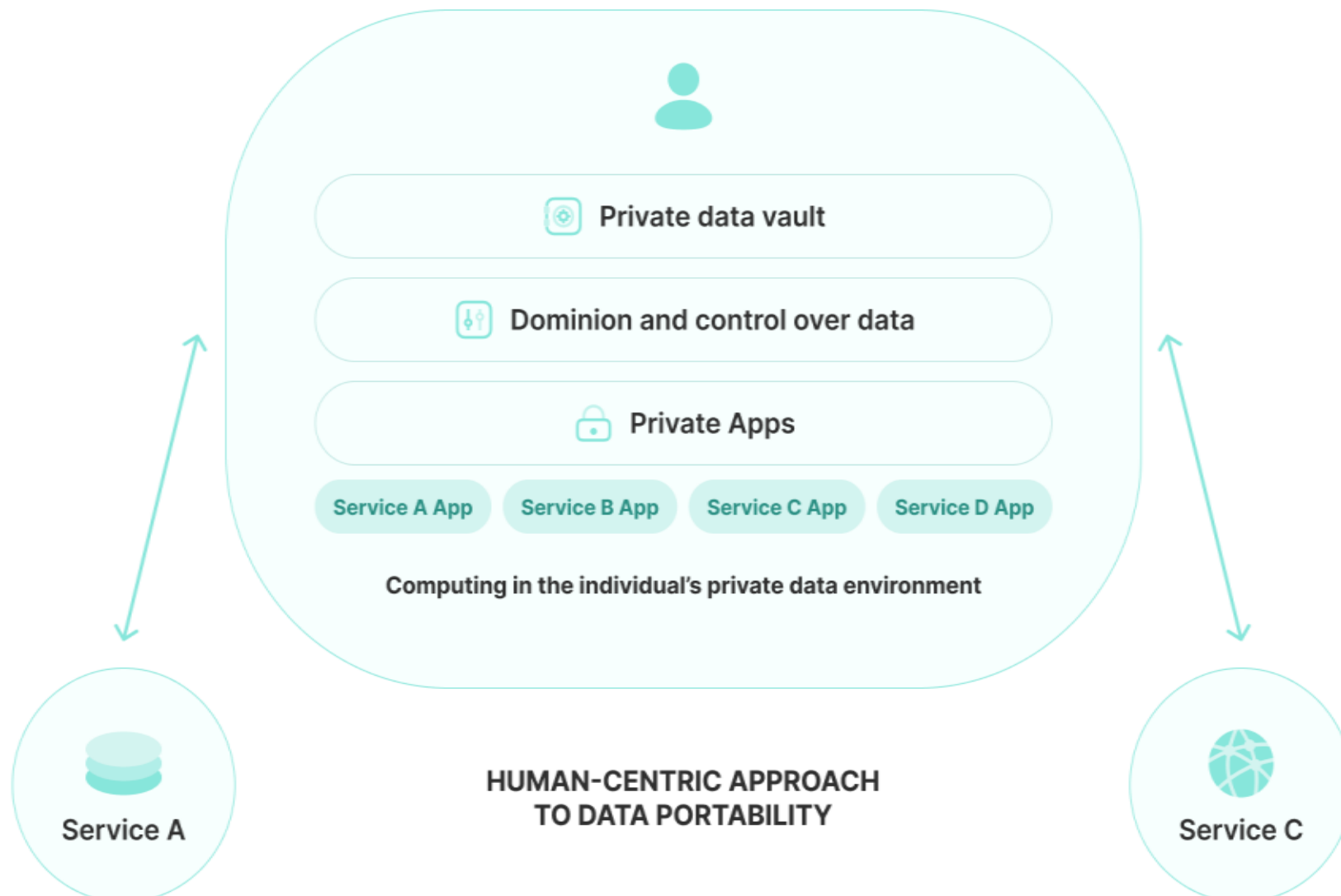

One of the fundamental features of this human-centric data ecosystem is that the user's data is private-by-default; only the user has access to their data in their personal data environments.[248] If a third-party app needs access to an individual's data (like location data or payment history), the application must obtain prior permission from the user to access such data and justify why access is necessary.[249]

Compared with the prior two data models, the human-centric data infrastructure rests on the assumption that an individual's data is a unique economic asset and a modern form of personal luxury. Moreover, user-generated data across platforms and services constitute an essential part of each individual's digital identity.[250] As a result, a human-centric approach embodies the idea that individuals should have dominion and control over their own data; it also embodies the principle that user-generated data should remain on the individual user's side.[251]

As such, a human-centric data model opens new opportunities with personal data. Developers and individuals can easily build

248. *Id.*

249. *See id.*; Paulius Jurcys, Chris Donewald, Jure Globocnik & Markus Lampinen, *My Data, My Terms: A Proposal for Personal Data Use Licences*, 33 HARV. J.L. & TECH DIG. 1, 13 (2020).

250. Jurcys et al., *supra* note 12, at 14.

251. Fenwick & Jurcys, *supra* note 9, at 391; *see* Jurcys et al., *supra* note 12, at 18.

applications without having to solve complex data and back-end problems.[252] Federated data architecture, where users own and control their personal data, frees developers from the hefty burden of complying with data privacy regulations.[253] This is because apps can operate locally on top of user-held data and generate value on the user side instead of being centralized on the service provider's platform.[254] Instead of solving complex data access, data formatting, and back-end issues, in the human-centric data ecosystem, third-party developers can build AI-powered apps that mostly consist of front-end solutions plugged in via APIs to the user's own data environment.[255] This is possible because of the separation of data from the apps. Furthermore, the human-centric approach to data allows developers to deploy personal AI assistants and agents in a private and secure environment, with access to rich, up-to-date, and holistic personal data of each individual.[256]

Comparing the enterprise and service-centric data governance models with a human-centric approach, the latter allows for a better understanding of how the power of data silos and information intermediaries can change. Most obviously, data portability is reconceptualized. In the human-centric, user-held model, the single point of focus is the individual.[257] Because each individual has different and unique data sets about themselves, it makes sense that AI-powered applications should be deployed on top of each individual's data—especially when it comes to highly sensitive, personal data (e.g., data from different sensors, wearable devices, and applications that contain such highly personal biodata or behavioral data).

From the perspective of dominion and control, a human-centric approach provides an even more comprehensive solution for personal data management than the data pod model alone. It allows individuals to fully control their personal data and private cloud computing resources, and use them for private applications and services, like personal AI and machine learning models.[258] As such, a user-centric data architecture makes it possible for individuals to run personal AI models in a private and secure environment, with access to rich, up-to-date, and holistic personal data.

---

252. . Fenwick & Jurcys, *supra* note 9, at 391. This is because the data is already cleaned and organized in each individual's personal data environment. *See id.*

253. *See id.*

254. *Id.*

255. *Id.* In academic literature, this model has sometimes been described as "*in situ*" data portability, which "bring[s] the algorithms to the data [in situ] instead of bringing the data to the algorithms [*ex situ*]." *See* Marshall W. Van Alstyne, Georgios Petropoulos, Geoffrey Parker & Bertin Martens, *'In Situ' Data Rights*, 64 COMMUN. ACM 34, 35, (2021).

256. *See* Fenwick & Jurcys, *supra* note 9, at 391.

257. *See id.*

258. *See id.*

## IV. PORTABILITY REIMAGINED: FROM ACCESS TO DOMINION AND CONTROL

Based on the emerging technical and market-driven data frameworks described in Part III, this Article offers a revisited concept of data portability and highlights some forgotten or neglected aspects of this concept, at least as articulated in current regulation. A more expansive concept of data portability points toward an alternative framework for thinking about data portability as a fundamental right and essential freedom. Such a concept provides a normative impetus for the technological architecture described above that offers individuals greater agency and control over their personal data.

### *A. The Essence of Portability*

The word portability is typically deployed as an adjective referring to something that is moveable; an object possesses the quality of portability if it can be easily carried or moved.[259] However, if we consider the meaning of this word more carefully, some nuances and complexities reveal and nudge us toward some more interesting possibilities. In particular, understandings of portability go beyond the simple conception of the moveable and spill out into a network of interconnected ideas and concepts. This becomes apparent when one considers the enormous number of words of which portability forms an element or with which it is connected.[260]

Implicit in the basic concept of portability is the notion of the transport or carriage of something moveable, impermanent, and personally valuable; it must travel securely and safely across space and

259. *Portability*, DICTIONARY.COM, https://www.dictionary.com/browse/portability [https://perma.cc/YTC4-KPU7] (last visited Jan. 31, 2025). This is not surprising, as the English word portability derives from the Latin word for carrying, *portare* (i.e., something that is capable of being transported or carried from place to place). *See Portare*, LATIN-DICTIONARY.COM, https://www.latin-dictionary.net/search/latin/portare [https://perma.cc/VBG8-ZG5L] (last visited Jan. 31, 2025). This usage can be seen carried forward into Late Latin, *portabilis*, which is defined as "that which can be carried," and the French word *portable* from which the English term derives. *See Portable*, ONLINE ETYMOLOGY DICTIONARY, https://www.etymonline.com/word/portable [https://perma.cc/5T9T-HEG4] (last visited Jan. 31, 2025).

260. For a partial list, consider comport; deport; disport; emporium; import; important; importune; opportune; opportunity; passport; porta (gate); harbor, port; portal; portcullis; porter; portfolio; portico; portiere; purport; practical; rapport; report; sport; support; transport; report; support; and important.

over time. In all portability, there is an implicit claim that individuals want to bring something of value to them into the future.

As such, portability relates to notions of individual freedom, autonomy, and empowerment. The projects an individual pursues in their life will be better if data is in their possession. Portability thus connects with a mentality or a state of independence from some constraint and the possibilities of a better future; it means an individual is liberated from being tied down to one place if they keep their data with them.

This line of thinking suggests a shift from portability as an adjective or quality of things towards portability as a more richly textured action that opens new possibilities and life projects. Crucially, that which is portable is both moveable and temporary but also important. It is the importance of the object that is portable. This reveals an interesting feature or paradox of portability; namely, that the impermanent—the moveable—is rendered central and essential. The essence of portability is this constitutive, essential impermanence.

Accordingly, portability reveals something important about human beings: we surround ourselves with moveable things that matter to us, and which can be thought of as important or essential for our identities. It is crucial that those things accompany us through time and space; we must enjoy the benefits of having them accompany us to sustain our identities. An essential part of our (digital) identity is our capacity to imagine new and better futures: plans, projects, and a striving for a better life in which realizing these better futures involves work and effort. The option of portability helps us realize these goals.

As such, portability is a powerful, alluring, all-encompassing notion that goes beyond the simple sense of an adjectival quality. Portability is something that is moveable towards an invocation of that which is essential to people and for people. Portability, in its essence, encompasses the concept of freedom. It refers to the capacity of an object, idea, or system to be easily transported, carried, or transferred from one place to another without significant constraint or alteration. But while the concept of portability is commonly associated with physical objects, its philosophical implications extend far beyond the realm of materiality.

At its core, portability embodies the human desire for flexibility and the ability to transcend boundaries. It represents our longing for liberation from the limitations imposed by space, time, and circumstance. By enabling portability, in a digital context, individuals seek to break free from the constraints that confine them, allowing them to explore new territories, engage in diverse experiences, and connect with people and ideas beyond their immediate surroundings.

From a philosophical perspective, portability can also be interpreted as a metaphorical concept. It speaks to an innate longing for personal growth, adaptability, and self-transformation. Just as physical objects can be transported, our minds, emotions, and perspectives can also traverse different landscapes and evolve. Portability encourages us to expand our horizons, challenge our beliefs, and embrace new ideas. It prompts us to question the boundaries we set for ourselves and explore alternative paths of thinking and being.

The meaning of portability encompasses a wide array of philosophical dimensions. It symbolizes society's yearning for freedom, flexibility, and connection. It encompasses the ability to transcend physical and metaphorical space and the creation of the new. Portability invites us to reflect on our transient nature and embrace the ever-changing nature of existence. Ultimately, it prompts us to recognize that our journey is not defined by the places we visit or the objects we carry, but by the profound experiences, connections, and insights we gather along the way.

### *B. Regulating Data Portability in the AI-Driven Ecosystems*

It is this richer and more nuanced meaning of portability that should be reflected in the legal deployment of portability in the context of data governance. The metaquestion is whether the contemporary legal discourse around data portability is sufficiently cognizant of this nuanced meaning. The current legal usage of portability captures something important about our relationship with our data—it emphasizes its moveability, its enormous value to each individual, and the individual's desire to control when and where it is moved.[261] It also embodies individuals' desires to use their data in different contexts, without red tape and with their consent. Individuals who want to use their data in other contexts and different digital environments should be able to. This idea has intuitive appeal and force—it also emphasizes the idea of individuals engaged in meaningful life projects and the importance of individual-owned data use in those life projects. There is a carrier—either the individual or some third-party data handler—and a means of carriage, like a public or private data network.

But does the current legal framework go far enough in recognizing the central importance of portability? Does it recognize the essential role and character of personal data in our lives and to our identity in a digital age? Legal concepts may frame an issue in a

261. See discussion, *supra* Sections III.C–IV.A.

particular way and inevitably have blind spots wherein aspects of an issue are obscured and attention is deflected away from other parts of the problem. The currently deployed legal concept of data portability obscures issues and limits the horizon of possibilities in at least three ways.

First, for individual consumers, data portability, as currently articulated and understood in the legal literature, is not a priority.[262] Consider a counterfactual thought experiment: is data portability, in a narrow sense, what individuals really want? That is, does the right to receive a copy of their data from a data controller or ask one data controller to move it to another data controller satisfy the desire for control over personal data? The answer is likely that one's own data is so important to the individual that they do not want a superqualified right of portability subject to the needs of service providers. Instead, the user likely seeks not only control, but full dominion over their data.[263] As such, data portability is not satisfying the interests of ordinary individuals, their desire to have control over their data, and their desire to leverage data in their life projects.

In a digital, AI-powered age, what matters more to a person's identity than their own data? This Article suggests that, in an increasingly technologically driven society, a person's data is increasingly constitutive of their personhood.[264] Data is personal in a deep, existential sense—it has become a defining feature of human identity and individual autonomy.[265] Indeed, it is hard to conceive of anything more central to identity in a digital age than personal data. As such, in the twenty-first century, our data must be recognized as inseparable from who—and what—we are. A more aggressive legal approach in which portability is conceptualized as a fundamental rather than a qualified right is needed. The EU AI Act's ambition to introduce a human-centric approach to AI technologies seems to lean that way, although vaguely.[266]

Data portability is presented as the empowering of individuals, yet it typically serves the interests of service providers.[267] Data portability understood in GDPR-like terms of sharing personal data to other services and between services has use cases, to be sure.[268] Still, there is a feeling that most, if not all, GDPR-like use cases are primarily designed from the perspective of adding value for services first. These portability solutions provide a framework for the deeply rooted digital services ecosystem. As such, current notions of data portability confuse the interests that are being served.

---

262. *See* discussion, *supra* note 4.

263. *See generally* Winegar & Sunstein, *supra* note 2 (studying the value of personal data, empirically proving the existence of the superendowment effect with regard to personal data).

264. *See id.*; Jurcys et al. *supra* note 12 (discussing the value of personal data in the age of personal AI agents and twins).

265. *See, e.g.*, Bertin Mertens, *Data Access, Consumer Interests and Social Welfare – An Economic Perspective on Data*, *in* GERMAN FEDERAL MINISTRY OF JUSTICE, *supra* note 13, at 7; *see also* Drexl, *supra* note 13, at 485.

266. *See* discussion, *supra* Section II.E.

267. *See* discussion, *supra* Section 3.2

Human-centric technology raises the tantalizing possibility that individuals can have dominion and control over their own data. The current technological environments, where data flow depends on siloed API frameworks, will likely be disrupted by the rapid evolution of natural language models, machine learning, and AI.[269] This is where a human-centric approach to personal data necessitates a complete reshaping of data portability. Instead of relying on the right to "port" fragmented portions of user data from Service Provider A to Service Provider B (the consent and opt-out model), individuals should be empowered to carry their data with them while service providers come to the individual.[270] As new, human-centric digital ecosystems emerge, they are also likely to provide an alternative to the currently prevailing "tracked-by-default" mechanisms, which place the burden on individuals to opt out.[271] In this new human-centric data paradigm, an individual's data remains on the user side, is private by default, and users decide which uses of their data they want to opt into.

## V. PATHS FORWARD

Our data is no longer peripheral—it is integral to individuals' existence in a digital age. As the world witnesses the gradual emergence of a human-centric approach to data portability, it becomes evident that personal data is not something separate from individuals but rather an intrinsic part of who and what they are. It defines individuals' digital personhood and shapes their interactions in the modern world. Therefore, it is imperative that the society acknowledges such a constitutive role of data and data portability in individuals' digital lives. This Article has shown that, in order to protect individual autonomy and maintain control over digital identities, society must adopt a more aggressive legal approach that recognizes data portability as an inalienable fundamental right, not merely a qualified right to opt out and manually porting data from one service to another. Only by doing

268. *See* discussion, *supra* Section 3.1.

269. *See* discussion, *supra* Section 1.

270. *See* discussion, *supra* Sections 3.3–4.2.

271. *See* discussion, *supra* Section 3.3.

so can individuals ensure that their data remains with them, always, as an essential component of their digital selves.

* * *

If you have any questions related to issues discussed in this paper, you can ask them to digital knowledge twins of Paul Jurcys[272] or Mark Fenwick:[273]

Talk to "Paul AI"

Talk to "Mark AI"

**Funding disclosures**

1) Timo Minssen's research for this paper was supported by a Novo Nordisk Foundation Grant for a scientifically independent International Collaborative Bioscience Innovation & Law Programme (Inter-CeBIL programme - grant no. NNF23SA0087056). Timo Minssen's research for this paper was further funded by the European Union (Grant Agreement no. 101057321; the "CLASSICA project"). Views and opinions expressed are however those of the author(s) only and do not necessarily reflect those of the European Union or the Health and Digital Executive Agency. Neither the European Union nor the granting authority can be held responsible for them.
2) Paulius Jurcys's research for this paper was supported by the JST Moonshot R&D Grant Number JPMJMS2215.

272. http://hey.speak-to.ai/paul.
273. https://hey.speak-to.ai/mark.